\documentclass[review]{elsarticle}

\usepackage{lineno,hyperref}
\usepackage{amsmath,amsfonts}
\usepackage{diagbox}
\usepackage{booktabs}
\usepackage{bm}  
\usepackage{algpseudocode}
\usepackage{algorithm}
\makeatletter
\newcommand{\removelatexerror}{\let\@latex@error\@gobble}
\makeatother
\usepackage{array}
\usepackage[subrefformat=parens,skip=3pt]{subcaption}
\usepackage{textcomp}
\usepackage{stfloats}
\usepackage{url}
\usepackage{amssymb}
\usepackage{placeins}
\usepackage{verbatim}
\usepackage{graphicx}
\usepackage{multicol}
\usepackage{multirow}
\usepackage{color}
\usepackage{float}
\usepackage{pifont}
\newtheorem{definition}{Definition}
\newtheorem{theorem}{Theorem}
\newtheorem{Lemma}{Lemma}

\newtheorem{corollary}{Corollary}

\journal{Pattern Recognition}

\begin{document}

\begin{frontmatter}

\title{Dynamic Proxy Domain Generalizes the Crowd Localization by Better Binary Segmentation}

\author[1,2]{Junyu Gao}
\author[1,2]{Da Zhang}
\author[3]{Qiyu Wang}
\author[2]{Zhiyuan Zhao}
\author[2,1]{Xuelong Li\corref{cor1}}
\ead{xuelong\_li@ieee.org}
\cortext[cor1]{Corresponding author}
\affiliation[1]{
            organization={School of Artificial Intelligence, Optics and Electronics (iOPEN), Northwestern Polytechnical University},
            city={Xi'an},
            country={China}}
\affiliation[2]{
            organization={Institute of Artificial Intelligence (TeleAI), China Telecom},
            country={China}}
\affiliation[3]{
            organization={School Of Electronics And Information, Northwestern Polytechnical University},
            city={Xi'an},
            country={China}}





\begin{abstract}
Crowd localization aims to predict the precise location of each instance within an image. Current advanced methods utilize pixel-wise binary classification to address the congested prediction, 
where pixel-level thresholds convert prediction confidence into binary values for identifying pedestrian heads. Due to the extremely variable contents, counts, and scales in crowd scenes, the confidence-threshold learner is fragile and lacks generalization when encountering domain shifts. Moreover, in most cases, the target domain is unknown during training. Therefore, it is crucial to explore how to enhance the generalization of the confidence-threshold locator to latent target domains. In this paper, we propose a Dynamic Proxy Domain (DPD) method to improve the generalization of the learner under domain shifts. Concretely, informed by the theoretical analysis of the upper bound of generalization error risk for a binary classifier on latent target domains, we introduce a generated proxy domain to facilitate generalization. Then, based on this theory, we design a DPD algorithm consisting of a training paradigm and a proxy domain generator to enhance the domain generalization of the confidence-threshold learner. Additionally, we apply our method to five types of domain shift scenarios, demonstrating its effectiveness in generalizing crowd localization. Our code is available at \href{https://github.com/zhangda1018/DPD}{\textcolor{blue}{DPD}}.
\end{abstract}

\begin{keyword}
Dynamic Proxy Domain \sep Crowd Localization \sep Domain Adaptation \sep Binary Segmentation.
\end{keyword}

\end{frontmatter}


\section{Introduction}
\label{sec:intro}
Crowd localization aims to predict the precise location of each instance within an image \citep{gao2021domain}. 
Due to its wide range of potential applications, it has attracted significant attention from researchers, resulting in substantial success in fully supervised crowd localization \citep{liang2022focal}, facilitated by advanced pipelines \citep{zhu2023daot} and training paradigms \citep{xie2023striking}. 
Nevertheless, this impressive performance relies heavily on extensive annotated data, and the commonly adopted \textit{Empirical Risk Minimization} (ERM) assumes that the testing data is independently and identically distributed to the annotated data \citep{vapnik1991principles}. 
It is evident that this assumption is vulnerable when applied to data sampled from real crowd scenes, leading to significant performance deterioration when violated.
Furthermore, crowd scenes are not effectively recognized by the trained crowd locator during testing, indicating that the target domain distribution is agnostic during training \citep{zhang2022cross}.
Therefore, improving the generalization of crowd locators trained on a source domain to a latent target domain is crucial. 
This paper aims to stabilize crowd localization performance or enhance its generalization when encountering non-conforming data distributions, specifically addressing target agnostic domain generalization.

\begin{figure}
    \centering
    \includegraphics[width=0.7\textwidth]{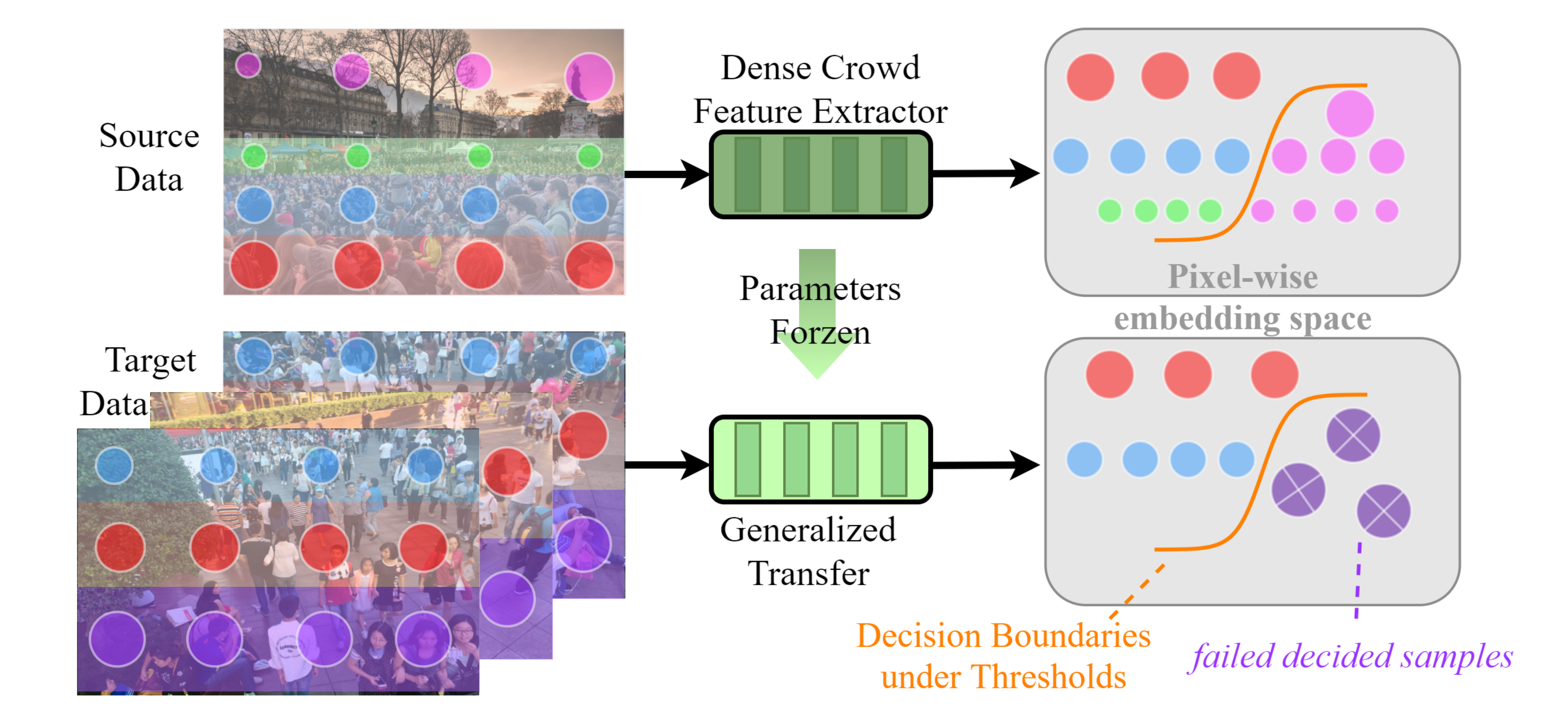}
    \caption{The superior performance achieved by existing segmentation based crowd locators mostly depends on the robust threshold to classify the samples into two parts. However, when transferring the threshold to another domain, the specific knowledge incurs some samples are ineffective under the thresholds.}
    \label{fig1}
\end{figure}

To begin with, we analyze the specific points for crowd localization under the domain shift issue. 
As aforementioned, the advanced pipelines lead to superior performance in crowd localization. 
For example, \citep{abousamra2021localization} proposes treating crowd localization as a binary segmentation task, where the head areas are segmented into foregrounds. 
However, due to variations in semantic knowledge about pedestrians (such as instance scale, exhibition, or scene style), the crowd locator exhibits varying confidence levels for instances within an image \citep{lin2023optimal}.
Therefore, \citep{gao2020learning} introduces a novel adaptive pixel-wise threshold learner to achieve variance-aware pixel-wise binary classification based on the extracted features. 
The fully supervised training paradigm then minimizes the empirical risk on the training set (or source domain), meaning the threshold learner aims to reduce empirical loss along with the non-convex loss landscape on annotated data in the source domain.

Unfortunately, crowd scenes are inherently subject to significant variations across images and datasets due to uncertainties in crowd features, such as scene layout, crowd count, and camera perspective, among others \citep{gao2022congested}. 
This leads to challenges in handling unseen scenes, known as the domain shift problem \citep{wang2021neuron}.
In such cases, the crowd locator often exhibits low confidence in object localization while showing excessive confidence in background areas. An illustrative example is presented in Fig. \ref{fig1}.
Additionally, incorrectly embedded features can lead to irrational adaptive thresholds \citep{gao2020learning}. 
Consequently, when the ERM process in the source domain rigidly directs the confidence threshold learner to the fixed source distribution, any domain shift exacerbates the difficulty of achieving effective generalization to the target domain. 
Paradoxically, pushing the model to overfit on the source distribution through ERM enhances source knowledge while distancing target-specific knowledge, a phenomenon known as the \textit{Matthew Effect}.
To this end, balancing confidence with thresholds is key to achieving crowd localization under domain generalization.

Based on the above observations, we propose a domain generalization framework for crowd localization, called \textbf{D}ynamic \textbf{P}roxy \textbf{D}omain (DPD), which is an attempt based on analyzing the upper bound of the generalization error in the target domain.
Specifically, we treat the confidence-threshold learner as a binary classifier. 
Then, by theoretically analyzing the generalization error upper bound in the target domain, we propose to generalize the source domain-trained model by introducing a new dynamic domain as a proxy.
Furthermore, ERM is able to push the model towards the dynamic domain distribution, rather than the fixed source one, making it feasible to enhance generalization in the target domain. 
According to the exploited theoretical guarantees, we design the corresponding algorithm, which is composed of source samples, a proxy domain generator, and a convergence strategy. 
In summary, the contributions of our work are threefold:

\begin{itemize}
    \item Propose to tackle the domain generalization of crowd localization from the perspective of generalizing the confidence-threshold learner. To the best of our knowledge, this paper is the first attempt on the issue.
    \item Present that a dynamic proxy domain generated from source-only data improves the generalization for binary segmentation based crowd locator while providing rigorous theoretical guarantees.
    \item Based on the theory, we design an algorithm for introducing dynamic proxy domain and its corresponding training paradigm and conduct experiments to provide empirical guarantees.
\end{itemize}

\section{Related Work}
\label{section2}

\subsection{Crowd Analysis}
 
The existing crowd analysis involves counting and localization (detection) \citep{goel2024learning}. 
Crowd counting has developed significantly due to its succinct but effective framework \citep{liu2024consistency}. 
Moreover, some studies extend it into more fields, such as multi-modal \citep{zhou2023mc}, multi-view \citep{zhang2025mahalanobis}, un-/semi-/weakly/noisy \citep{liang2023crowdclip} supervised learning. 
Crowd localization also attracts research attention as it offers more information than counting. 
The purpose of crowd localization is to locate the exact position of each head in a scenario. 
Earlier locators are initially based on object detection \citep{redmon2016you}. 
Subsequently, researchers have extended work on addressing intrinsic scale shifts, but detection-based methods still perform poorly in extremely congested situations \citep{gao2022congested}. 
TinyFaces \citep{hu2017finding} uses a detection-based framework to locate tiny faces by analyzing the effects of scale, contextual semantic information and image resolution. 
Following this, some researchers have extended work on addressing intrinsic scale shifts \citep{li2019pyramidbox++}, yet detection-based methods continue to perform poorly in extremely congested situations.
Additionally, points-based locators \citep{li2022video} have been proposed. Li et al. \citep{li2022video} proposed a multi-focus Gaussian neighborhood attention to estimate exact locations of human heads in crowded videos. 
Although these methods worked to some extend, they cannot provide scale information, and performance is still undesirable. 
Thus, the pixel-wise binary segmentation \citep{abousamra2021localization, gao2020learning} is proposed for crowd localization. 
However, the training of thresholds suffer from overfitting on the training data (source domain). To generalize it to the target agnostic domains, we propose DPD, which enhances the model's adaptability and robustness within unknown target domains through the introduction of a dynamically generated proxy domain that simulates and adapts to diverse data distributions.

\subsection{Cross Domain Convergence}
 
Existing machine learning methods rely on training with large amounts of data. 
Specifically, the domain shift between training and testing data impedes the generalization of models. 
Hence, Wen et al. \citep{wen2019exploiting} first proposed enhancing performance on the target domain by introducing some unlabeled target data, a process known as \textit{Domain Adaptation} (DA)
Subsequently, several DA methods have been proposed \citep{wang2021neuron}. 
In DA, the traditional paradigms include adversarial training, self-training and few shot learning \citep{chen2024one}. 
Several methods attempt to adapt domains by finding their similarities \citep{zhu2022fine} while others try to discover common knowledge between them \citep{du2023domain}.
However, most of the time, the target domain is completely agnostic to us in training, which is addressed by \textit{Domain Generalization} (DG) \citep{xie2023striking}. 
Since there are only training samples available and we do not know how the target domain is distributed, the goal mainly focuses on enhancing generalization and reducing overfitting to the source domain \citep{peng2024single}.  
In this paper, our DPD achieves the two purposes via converging on source domain and dynamic proxy domain simultaneously. 
This method not only enhances the model's generalization ability in the known source domain but also simulates the characteristics of the target domain through a dynamically generated proxy domain, thereby improving the model's adaptability to unknown target domains without explicit information about them.

\section{Preliminary}
\subsection{Supervised Crowd Localization}

\begin{figure}[hbpt]
    \centering
    \includegraphics[width=0.6\textwidth]{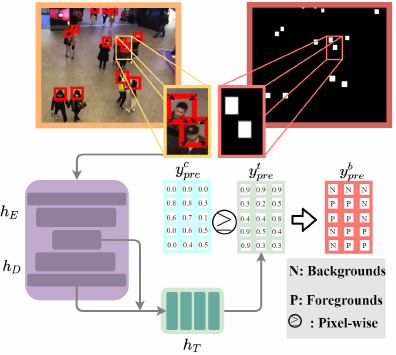}
    \caption{An example for the pipeline of adaptive instance crowd localization. To facilitate visualization, only the image patch in yellow window are fed into crowd locator.}
    \label{fig2}
\end{figure}

In crowd localization \citep{abousamra2021localization}, given a crowd image represented by $x\in \mathbb{R} ^{3\times H\times W}$, the encoder $h_E$ in the locator maps the image to a latent feature $h_{E}(x)$, which has a higher channel but lower resolution than the original image. The decoder $h_D$ utilizes this latent feature to generate a confidence map
$y_{pre}^c \in \mathbb{R} ^{1\times H\times W}$, where the confidence values indicate the likelihood of a given pixel being in pedestrians' head area. Then, a fixed threshold of 0.5 (due to binary segmentation) divides the confidence values into two categories and generates a binary map as shown in Fig. \ref{fig2}. Ideally, the relationship between $y_{pre}^c$ and the ground truth binary map of $y_{gt}^b$ should be formulated as shown in Eq. \ref{eq1}:
\begin{equation}
\left\{
\begin{aligned}
& \lim_{{y^{b}_{gt}(i,j)}=1} y^{c}_{pre}(i,j) \rightarrow 1^- & , & \quad (i,j) \in (H,W); \\
& \lim_{{y^{b}_{gt}(i,j)}=0} y^{c}_{pre}(i,j) \rightarrow 0^+ & , & \quad (i,j) \in (H,W),
\end{aligned}
\right.
\label{eq1}
\end{equation}
where a fixed threshold effectively separates foreground and backgrounds confidence. 
However, the instances among the crowds exhibit significant variance, resulting in very low predicted confidence values in $y_{pre}^c$ for some challenge or rare samples. For instance, the decoder $h_D$ struggles to consistently regress $y_{pre}^c(i,j)$ towards 1 in the head area and 0
in backgrounds, making it difficult to achieve accurate predictions. 
As a result, a fixed threshold fails to detect such instances.

To overcome this limitation and obtain precise predictions, \cite{gao2020learning} introduces a pixel-wise threshold map to adaptively separate $y_{pre}^c$ into a binary map $y_{pre}^b$. Hence, the locator is fed with an image $x\in \mathbb{R} ^{3\times H\times W}$ along with its corresponding binary map annotation $y\in N_{\{0,1\}}^{1\times H \times W}$. 
In order to derive an adaptive threshold map based on the latent feature $h_E(x)$, the threshold learner $h_T$ is proposed to map
$h_E(x)$ and $y_{pre}^c$ into a threshold map $y_{pre}^t \in \mathbb{R} ^{1\times H\times W}$. The learned threshold map $y_{pre}^t$ enables it to lower the threshold for hard instances that are predicted with lower confidence by $h_D$. This aids to produce a more robust binary map which can be estimated via Eq. \ref{eq2}:
\begin{equation}
    y^b_{pre}=\left \lceil y^c_{pre}\ge  y^t_{pre}  \right \rfloor ,
    \label{eq2}
\end{equation}
where the $\left \lceil  \right \rfloor  $ is the pixel-wise Iverson bracket. $y^c_{pre}$ and $y^ct_{pre}$ are from Eq. \ref{eq3}:
\begin{align}
    \notag y^t_{pre} & = h_T[h_E(x)*y^c_{pre}], \\
    y^c_{pre} & = \text{Sigmoid}\left \{ h_D[h_E(x)] \right \}.
    \label{eq3}
\end{align}
A visual representation of the process can be found in Fig. \ref{fig2}. To this end, according to the above arrayed process and the mapping function $h$ in the hypothetical annotation space $\mathcal{H}$ can be induced as Eq. \ref{eq4}, which means to find the best pixel-wise classification prediction $y_{pred}^b$ such that the difference from the actual pixel classifications $y^b_{gt}$ is minimized ($\oplus$ represents XOR operations at the pixel level: the same is 0, the difference is 1):
\begin{equation}
    h: x\mapsto arg\max_{y^b_{pre}} \sum_{i}^{H\times W} y^b_{pre}\oplus y^b_{gt}.
\label{eq4}
\end{equation}
However, \cite{gao2020learning} further enhances the training via adding an optimization term to $y^c_{pre}$. The \textit{empirical risk} of a given $h\in \mathcal{H}$ are formulated as Eq. \ref{ER},
\begin{align}
    \widehat{R}(h)&\triangleq\frac{1}{N}\sum_{i=1}^{N}  [\mathcal{L}_2(y^c_{pre}, y^b_{gt}) +\mathcal{L}_1(y^b_{pre}, y^b_{gt})],
    \label{ER}
\end{align}
in which the $\mathcal{L}_n(\cdot, \cdot)$ represents the norm $n$ loss function and $N$ is the number of samples. Hence, the \textit{Empirical Risk Minimization} (ERM) function is:
\begin{equation}
    ERM(h) = arg \min_{h \in \mathcal{H}} \widehat{R}(h).
    \label{ERM}
\end{equation}
To better clarify the pipeline of adaptive threshold crowd localization, a pseudo code is arrayed in the Appendix \hyperref[pseudo1]{A.1}.

However, when the testing data does not obey independent identically distributed (\textit{i.i.d.}), namely domain generalization issue, after training with Eq. \ref{ERM} the distribution of $\text{Pr}(y^c_{pre})$ and $\text{Pr}(y^t_{pre})$ tend to be irrational, as shown in Eq. \ref{eq7}, which is opposite to Eq. \ref{eq2}.
\begin{equation}
\left\{
\begin{array}{ll}
y^{c}_{pre} \geq y^{t}_{pre}, & y^b_{gt} = 0 \\
y^{c}_{pre} < y^{t}_{pre}, & y^b_{gt} = 1
\end{array}
\right.
\label{eq7}
\end{equation}
Based on these issues, how to derive a domain-robust threshold-confidence learner and repair the irrational pairs in Eq. \ref{eq7}? We will firstly provide some theoretical preliminaries on the irrationality under domain generalization.

\subsection{Theoretical Analysis on Cross Domain Convergence}
Let $\mathcal{D}_s$ be the set of source domain, which is a distribution involving input crowd sample space $\mathcal{X}_s$ along with its ground truth annotations space $\mathcal{Y}_s$. Then, another domain $\mathcal{D}_t$ is introduced as the target distribution, which is defined as $\mathcal{X}_t\times\mathcal{Y}_t$. In practice, the \textit{domain generalization} task is fed by an \textit{i.i.d.} source sample drawn from $\mathcal{D}_s$ as the Eq. \ref{DASet} shown,
\begin{equation}
    \left \{ (x^s_i,y^s_i) \right \}_{i = 1}\sim \mathcal{D}_s.
    \label{DASet}
\end{equation}
Next, since $h$ is the mapping function of binary classifier, the \textit{error risk} on target space $\mathcal{D}_t$ is as Eq. \ref{target risk}:
\begin{equation}
    R_{\mathcal{X}_t \times  \mathcal{Y}_t  }\triangleq \mathrm{Pr} _{(x_t,y_t)\sim \mathcal{D}_t }\{\overrightarrow{\delta}  \lceil  h(x_t)\ne y_t \rfloor \},
    \label{target risk}
\end{equation}
where $\overrightarrow{\delta} $ is the two dimensional \textit{Dirac} function.

In domain generalization, the pain point is that the optimization objective is on target samples, while the real training is done on source samples. To this end, the discrepancy between distributions incurs model with low error risk on source domain hard to be also generalized well on target domain. That is, it's difficult to balance the discrepancy between Eq. \ref{ERM} and minimizing Eq. \ref{target risk}. More specifically, the discrepancy between distributions is the key. Hence, the former researchers \citep{ben2006analysis} leveraged $\mathcal{H}\bigtriangleup \mathcal{H}$-divergence to measure the discrepancy:
\begin{definition}\label{def-div}
    Let $\mathcal{D}_s$  and $\mathcal{D}_t$ be the two aforementioned domains distribution, $h$ is the hypothesis to the mapping function, while $h_s$ is the converged one. The $\mathcal{H}\bigtriangleup \mathcal{H}$-divergence between source and target is
    \begin{equation}    
        div_{\mathcal{H}\bigtriangleup \mathcal{H}}=\sup_{h,h_s\in\mathcal{H} }^{} \left | \mathbb{E} _\mathcal{S} \lceil h_s \ne h   \rfloor - \mathbb{E} _\mathcal{T} \lceil h_s\ne h \rfloor \right |_1 .
    \end{equation}
\end{definition}
\noindent
However, given a sampled data set from distribution, the Def. \ref{def-div} is limited and hard to be computed. Thus, \citep{ben2006analysis} approximate it  by introducing Def. \ref{def-PVT} via a proxy divergence:
\begin{definition}\label{def-PVT}
    A proxy dataset is constructed as:
        \begin{equation}
            \mathcal{X}_{prox}=\left \{ (x_i,\lceil x_i\sim \mathcal{D}_s\rfloor) |i\in \{0,\cdots ,N_s+N_t\}\right \} .
        \end{equation}
    A proxy generalized error $\epsilon_p$ is introduced on $\mathcal{X}_{prox}$. Then, using $\mathcal{A}-$distance ($A$ is some specific part of $\mathcal{X}_{prox}$ and $\mathcal{A}$ is the set of them), the $\mathcal{H}\bigtriangleup  \mathcal{H}\text{-}$divergence can be approximated as:
    \begin{equation}
        \hat{div}_{\mathcal{H}\bigtriangleup \mathcal{H}}=2\cdot (1-2\epsilon_p)=2\sup_{A\in \mathcal{A} } \left | \operatorname{Pr}_{\mathcal{D}_s}(A)- \operatorname{Pr}_{\mathcal{D}_t}(A) \right |.
        \label{TVD}
    \end{equation}
\end{definition}
\noindent
Given the discrepancy between two domains, we are ready to measure the empirical risk on target domain under the cross domain settings. More specifically, the upper bound on the target error risk can be formulated as Lem. \ref{lemma-bound}, which is proposed and expanded:
\begin{Lemma}\label{lemma-bound}
    Assume that the $\mathcal{H}$ is a hypothesis space with a VC dimension of $d$ and $m$ is the number of training samples, drawn from $\mathcal{D}_s$. Given an $h \in \mathcal{H}$, which is a binary classifier, the following inequality holds (specific proof is in Appendix 
    \hyperref[proof_lemma1]{B.1}.) with a probability at least $1 - \delta $, where $\delta \in (0, 1)$:
    \begin{align}\scriptsize
        \notag R_{\mathcal{T}}(h)&\le {R}_{\mathcal{S}}(h)+\frac{1}{2} \hat{div}_{\mathcal{H}\bigtriangleup\mathcal{H} }( \mathcal{D}_{s}, \mathcal{D}_{t})  \\
        &+4\sqrt{\frac{2dlog(2m)+log(\frac{2}{\delta})}{m} }+\lambda,
        \label{bound}
    \end{align}
    in which 
    \begin{equation}
        \lambda=\inf_{\hat{h}\in \mathcal{H}}^{}\left [R_{\mathcal{S}}({\hat{h}}) + R_{\mathcal{T}}(\hat{h})\right ].
    \end{equation}
\end{Lemma}
\noindent
Therefore, a generalized model can be achieved with a tighter upper bound to the error risk in target domain, namely the right hand term in Eq. \ref{bound}.

\section{Method}
\label{section3}

\subsection{Theory of DPD}

\begin{figure*}[hbpt]
    \centering
    \includegraphics[width=0.95\textwidth]{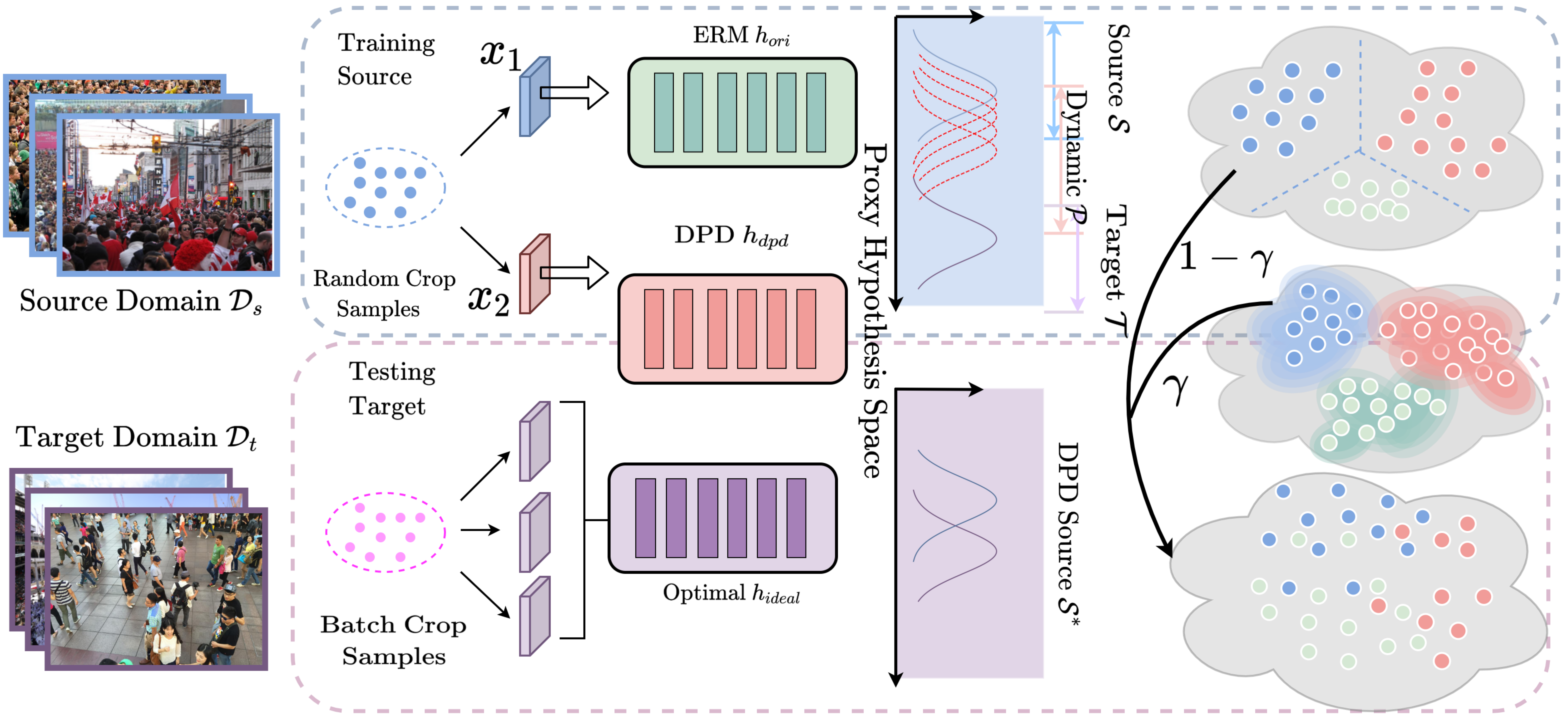}
    \caption{Overview of our core idea to the proposed \textbf{Dynamic Proxy Domain}. Comparing with implementing ERM on the source domain, we introduce DPD which minimizes the divergence between source distribution with target distribution. To this end, the decision boundary among domains can be weaken on the hypothesis space.}
    \label{fig3}
\end{figure*}

In conventional cross domain scenarios, only two domains exist, namely source $\mathcal{D}_{s}$ and target $\mathcal{D}_{t}$. As for domain generalization, directly training on $\mathcal{D}_{s}$ then testing on $\mathcal{D}_{t}$ results in poor performance. To this end, we propose a new domain named Dynamic Proxy Domain $\mathcal{D}_{p}$.

\begin{definition}\label{DPD-Def}
Given source distribution of $D_{s}$ and target distribution of $\mathcal{D}_{t}$, an additional $\mathcal{D}_{p}$ with Eq. \ref{DPD} holding,
\begin{equation}
    \operatorname{div}_{\mathcal{H}\bigtriangleup \mathcal{H}}\left(\mathcal{D}_{p}, \mathcal{D}_{t}\right) < \operatorname{div}_{\mathcal{H}\bigtriangleup \mathcal{H}}\left(\mathcal{D}_{s}, \mathcal{D}_{t}\right)
    \label{DPD}
\end{equation}
is called Dynamic Proxy Domain, which will be used in training to supplement $\mathcal{D}_{s}$.
\end{definition}

By introducing DPD (Fig. \ref{fig3}), we can derive a tighter upper bound to generalization error risk. Firstly, we need to derive specific formula to the upper bound to the state training with DPD. According to Def. \ref{DPD-Def}, the introduced DPD is in the training period, in which the model is fitting on $\mathcal{D}_{p}$ and $\mathcal{D}_{s}$ simultaneously. To this end, we put forward a theorem of Thm. \ref{thm:number}, which is the new upper bound to generalization error risk on target domain training with $\mathcal{D}_{s}$ and $\mathcal{D}_{p}$ simultaneously. The proof can be found in the Appendix \hyperref[proof_theorem1]{B.3}..

\begin{theorem}\label{thm:number}
 Let $h$ be the binary classifier hypothesis in the $\mathcal{H}$ with a VC-dimension of $d$ and $m_s, m_p$ are the number of source/proxy samples. Let $\mathcal{D}_{p}$ be the empirical distribution drawn \textit{i.i.d.} from the dynamic proxy domain. Then, a hyper-parameter  $\gamma \in [0,1]$ is defined, which is the convex combination rate. Thus, for any $\delta \in (0,1)$, with probability at least $1-\delta$,
\begin{align}\scriptsize
    \notag R_{\mathcal{T}}(h) &\leq \gamma \cdot \left( \hat{R}_{\mathcal{S}}(h) + \frac{1}{2} \operatorname{div}_{\mathcal{H}\bigtriangleup \mathcal{H}}\left( \mathcal{D}_{s}, \mathcal{D}_{t} \right) \right) \\
    \notag&+ (1-\gamma) \cdot \left( \hat{R}_{\mathcal{P}}(h) + \frac{1}{2} \operatorname{div}_{\mathcal{H}\bigtriangleup \mathcal{H}}\left( \mathcal{D}_{p}, \mathcal{D}_{t} \right) \right) \\
    &+ \lambda_{\gamma} + 4 \sqrt{ \frac{2d \log 2 \left( m_{s} + m_{p} \right) + \log \left( \frac{2}{\delta} \right)}{m_{s} + m_{p}} },
            \label{bound_another}
\end{align}
in which
\begin{align}
\lambda_{\gamma} &= \inf_{{h} \in \mathcal{H}} \left[ \gamma \cdot R_{\mathcal{S}}({\hat{h}}) + (1-\gamma) \cdot R_{\mathcal{P}}({\hat{h}}) + R_{\mathcal{T}}({\hat{h}}) \right].
\end{align} 
\end{theorem}

Let us compare the formula of upper bound of target error risk under naive ERM training (Lem. \ref{lemma-bound}) with DPD training (Thm. \ref{thm:number}). We prove that the introduction of DPD indeed provides an effect on training with the aim of influencing the upper bound to the target error risk. Thus, let us consider the case when introducing DPD incurs nothing on original training, namely the conditions for the establishment of the equal sign between the right-hand term in Eq. \ref{bound_another} with the right-hand term in Eq. \ref{bound}, which is shown by Cor. \ref{Coro-1}.

\begin{corollary} \label{Coro-1}
Let  $\Theta(h_{\text{DPD}}) = \Theta(h_{\text{ERM}})$  hold, where  $\Theta(\cdot)$  is the right-hand term of Eq. \ref{bound_another} with Eq. \ref{bound}. It is easy to derive  $\gamma = 1$ , considering Eq. \ref{DPD} provides condition on DPD. Moreover, the number of effective proxy training sample $m_p$ should degenerate to 0, which only holds under two cases: $\mathcal{D}_p$  did not involve in training or $\mathcal{D}_p = \mathcal{D}_s$ .
\end{corollary}

The Cor. \ref{Coro-1} tells us when $\gamma = 1$ , the Eq. \ref{bound_another} equals Eq. \ref{bound}. On the one hand, it is obvious that as long as DPD is introduced,  $\gamma < 1$ . On the other hand,  $\mathcal{D}_{p} = \mathcal{D}_{s}$  goes with obvious paradox with Definition to DPD. By now, we demonstrate the introduction of DPD indeed influences the upper bound. As for whether it deduces or enlarges the upper bound, let us decompose the $\Theta(h_{DPD})$ term by term.

In Thm. \ref{thm:number}, the terms $\widehat{R}_{\mathcal{S}}(h), \widehat{R}_{\mathcal{P}}(h)$ are empirical error risk on training domain, which can be very low as our training paradigm is based on ERM. For the last term namely  $\lambda$ , when the  $\lambda$  is large, it is impossible to generalize model to the $\mathcal{D}_{t}$ \citep{ben2010theory}. As for the term with root sign, it can be easily proved (see Appendix \hyperref[proof_number]{B.2}) by the monotonicity of the whole term with respect to  $m$ , namely a larger  $m$  (denotes  $m_{s} + m_{p}$) results in a smaller whole term. Finally, the left term is the divergence of the joint domains to target domain. We transfer the conclusion into Thm. \ref{thm2} and put the proof to the Appendix \hyperref[proof_theorem2]{B.4}. As shown in Thm. \ref{thm2}, it tells us the introduction of dynamic proxy domain facilitates deriving a tighter upper bound to the generalization error risk on target domain with a smaller divergence as one of attributes.

\begin{theorem}\label{thm2}

Let  $ h_{\text{DPD}} $  be the DPD hypothesis and  $ h_{\text{ERM}} $  be the Empirical Risk Minimization hypothesis on the space of  $ \mathcal{H} $  with a VC-dimension of  $ d $. Then, for any  $ \delta \in (0,1) $ , with probability at least  $ 1-\delta $ , the Eq. \ref{eq:thm2_1} can be derived,
\begin{equation}
    \sup_{h_{\text{DPD}} \in \mathcal{H}} R_{\mathcal{T}}(h_{\text{DPD}}) \leq \sup_{h_{\text{ERM}} \in \mathcal{H}} R_{\mathcal{T}}(h_{\text{ERM}}),
    \label{eq:thm2_1}
\end{equation}
with Eq. \ref{eq:thm2_2} as the necessary and insufficient condition:
\begin{align}\scriptsize
    \gamma \cdot \operatorname{div}_{\mathcal{H}\bigtriangleup \mathcal{H}}(\mathcal{D}_{s}, \mathcal{D}_{p})  + (1-\gamma) \cdot \operatorname{div}_{\mathcal{H}\bigtriangleup \mathcal{H}}(\mathcal{D}_{s}, \mathcal{D}_{p}) > \operatorname{div}_{\mathcal{H}\bigtriangleup \mathcal{H}}(\mathcal{D}_{s}, \mathcal{D}_{p}).
    \label{eq:thm2_2}
\end{align}
\end{theorem}

\subsection{Algorithm of DPD}

The theoretical guarantees proposed above will be used to design an algorithm for domain  generalization crowd localization using Dynamic Proxy Domain (DPD). Let us recall what a theoretical analysis chain DPD shows us.
Firstly, Def. \ref{DPD-Def} tells us what is DPD, along with a significant property of Eq. \ref{DPD}. Then, Thm. \ref{thm:number} derives the upper bound to the target error risk trained with DPD, and Cor. \ref{Coro-1} demonstrates its impact on generalization capacity for cross-domain crowd localization. 
Finally, Thm. \ref{thm2} provides conclusive statement that DPD can reduce the upper bound of generalization error risk on the target domain. With
the guarantee of theoretical analysis, our algorithm for DPD will follow these results.

To begin with, in Lem. \ref{lemma-bound}, the first term namely $\widehat{R}_{\mathcal{S}}(h)$ is empirical error risk on source domain, which can be very low as our training paradigm is based on ERM. 
As for the last term namely $\lambda$, when the $\lambda$ is large, it is impossible to generalize model to the $\mathcal{D}_t$. 
Finally, our optimization target lies on the two terms in the middle of Eq. \ref{bound}, which are $\varepsilon(m)$ and $div_{\mathcal{H}\bigtriangleup\mathcal{H}}(\mathcal{D}_{s}, \mathcal{D}_{t})$. 
That is, the generalization error risk on target domain $R_\mathcal{T}$ can be bounded from two terms, which are $\varepsilon(m)$ and $div_{\mathcal{H}\bigtriangleup\mathcal{H}}(\mathcal{D}_{s}, \mathcal{D}_{t})$. Moreover, we still need to consider the empirical risk. Thus, our objective function firstly can be:
\begin{equation}
    \min_{h\in\mathcal{H}}[\hat{R}_{\mathcal{S}}(h)+\varepsilon(m)+div_{\mathcal{H}\bigtriangleup\mathcal{H}}(\mathcal{D}_{s}, \mathcal{D}_{t})].
    \label{subject}
\end{equation}
\subsubsection{Momentum Network for Usage of Source Data}\label{sec:4-1}
Eq. \ref{subject} shows the three terms of our objective. In this subsection, we show how the DPD optimizes the first two terms. For the $\hat{R}_{\mathcal{S}}(h)$, it is a normal ERM process. Thus, given a batch of samples $\mathbf{x}\in\mathbb{R}^{B\times3\times H\times W} $, the hypothesis $h$ is able to map it into $\mathbf{y}^c_{pre} \in\mathbb{R}^{B\times1\times H\times W}$ and $\mathbf{y}^b_{pre}\in\mathbb{N}^{B\times1\times H\times W}_{\{0,1\}} $. For the ERM part, the optimization problem can be arrayed:
\begin{equation}
    \min_{h\in\mathcal{H}}(\left \| {y}^c_{pre}-{y}^b_{gt} \right \| ^2+\left \| \mathbf{y}^b_{pre}-{y}^b_{gt}\right \|^1 ).
    \label{suped}
\end{equation}
The Thm. \ref{thm:number} tells that a larger $m$ can effectively reduce $\varepsilon(m)$. However, due to batch manner training, we cannot introduce too many samples in one gradient descend (GD) step. 
To this end, we propose a \textit{Momentum Updated Model} $\mathcal{M}_{Mo}$ to equally achieve zooming $m$. Moreover, thanks to the usually adopted training paradigm in crowd localization, namely \textit{random  crop}, we can fully utilize it to further enhance the zooming. 

To be concrete, given two cropped source images, which are ${x}_1,{x}_2 \in\mathbb{R}^{B\times3\times H\times W} $, we utilize one of them namely ${x}_1$ to train with ERM through Eq. \ref{suped}. Then, another crop ${x}_2$ is predicted by $h$ and $\mathcal{M}_{Mo}$ simultaneously. Finally, a consistency constraint is introduced:
\begin{equation}
    \min_{h,\mathcal{M}_{Mo}\in\mathcal{H}}(\left \| \mathbf{y}^c_{pre}-\mathbf{y}_{Mo}^c \right \| ^2+\left \| \mathbf{y}^b_{pre}-\mathbf{y}^b_{Mo}\right \| ).
    \label{consis}
\end{equation}
In addition, the parameters of momentum model $\theta _{Mo}$ are updated as:
\begin{equation}
    \theta _{Mo}\longleftarrow \mu  \cdot \theta _{Mo}+ (1-\mu ) \cdot \theta _{h},
    \label{update}
\end{equation}
where $ \mu  \in[0,1]$ is the updating coefficient.
Through combining Eq. \ref{consis} with Eq. \ref{suped}, the ERM reduces the $\hat{R}_{\mathcal{S}}(h)$ on the one hand, and the number of training samples $m$ is zoomed on the other hand.

\subsubsection{Dynamic Proxy Domain}
\begin{figure}
    \centering
    \includegraphics[width=0.5\textwidth]{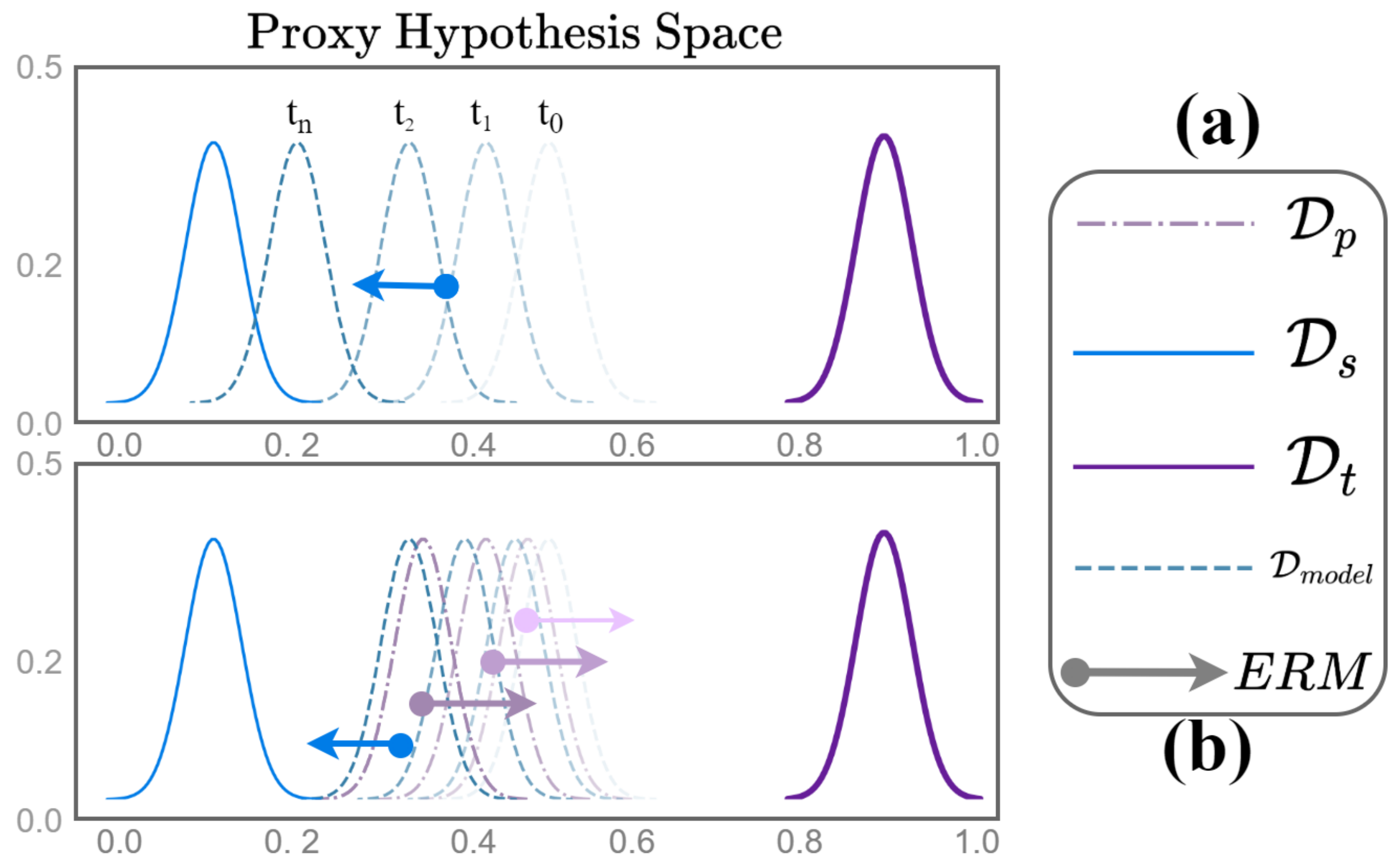}
    \caption{(a) Convergence to the fixed source domain. (b) Convergence to the source domain along with dynamic proxy domain simultaneously.}
    \label{DPD_Conv}
\end{figure}

In this subsection, we show how our proposed DPD is introduced. To begin with, we refer to the theoretical guarantees, in which a \textit{dynamic proxy domain} facilitates the generalization. Therefore, the key lies on the generation of the dynamic proxy domain $\mathcal{D}_p$. Firstly, the Lem. \ref{lemma-bound} tells us as the fitting degree being enhanced, the generalization is weaker as a result of the existence of ${d}_{\mathcal{H}\bigtriangleup  \mathcal{H}}(\mathcal{D}_{s}, \mathcal{D}_{t})$. However, since the source domain is all knowledge we have, there seems no other way to let crowd locator get the localization knowledge (how to embed the image into instance confidence and threshold). To this end, we notice that the parameters before overfitting on source domain could reserve more generalized knowledge (but less localization knowledge). Inspired by this, we propose that the generation of $\mathcal{D}_p$ could be based on the model prediction before overfitting. A toy example has been illustrated as Fig. \ref{DPD_Conv}. By now, we notice that the Eq. \ref{consis} also minimizes the risk on generated history domain via Momentum model, which is composed of history parameters. However, the Eq. \ref{update} suggests that the parameters of Momentum model is being pushed to the main model, which can be deemed as overfitted one. Thus, we propose to generate $\mathcal{D}_p$ in the second order.

To be specific, we propose a \textit{Dynamic $\mathcal{H}\bigtriangleup  \mathcal{H}$-generator} namely $h_T^{DPD}$, which is an independent threshold learner and also defined on the same $\mathcal{H}$ space. Concretely, the \textit{Dynamic Proxy Domain} is generated via $h_T^{DPD}$. Moreover, the $h_T^{DPD}$ has the independent optimizer to update. Then, we can concatenate the proposed $h_T^{DPD}$ into the Momentum model in Sec. \ref{sec:4-1}. 

To begin with, we pick one crop between $\{x_1, x_2\}$. Then, the corresponding $y^c_{pre}$ is fed into the original threshold learner $h_{T}$ along with $h_T^{DPD}$ simultaneously. By now, the $h_T^{DPD}$ is able to generate $\mathcal{D}_p$ predictions during dynamic training. Then, to reduce the $d_{\mathcal{H}\bigtriangleup  \mathcal{H}}(\mathcal{D}_p, \mathcal{D}_s)$, our objective can be:
\begin{equation}
    \min_{h, h_T^{DPD}\in\mathcal{H}}(\mathcal{L}[h(x_s),  h_T^{DPD}(x_s)]),
    \label{DPDobj}
\end{equation}
where $\mathcal{L}(\cdot, \cdot)$ denotes the loss function.

To better utilize the introduced DPD, we exploit a training strategy that in the convergence of Eq. \ref{DPDobj}, a \textit{stronger} loss should be implemented than the one utilized in the second term of Eq. \ref{suped}. In this paper, the concrete implementation of Eq. \ref{DPDobj} is ($\circ$ represents XOR operations: if one of them is 1, the result is 1):
\begin{equation}
    \small
    \mathcal{L}_T^{DPD} =1-\frac{2\cdot \left \| y^b_{pre}\circ y^b_{dpd}  \right \|^1   }{\left \|y^b_{pre}  \right \|^1+\left \|y^b_{dpd}  \right \|^1 }+\left \| y^b_{dpd}-y^b_{pre} \right \| ^1.
    \label{eq24}
\end{equation}

We provide pseudo code of DPD Algorithm in Appendix \hyperref[pseudo2]{A.2}.

\section{Experiments}

\subsection{Datasets}
In this paper, we conduct our DPD on six datasets, which are SHHA, SHHB \citep{zhang2016single}, QNRF \citep{idrees2018composition}, JHU \citep{sindagi2019pushing}, NWPU \citep{wang2020nwpu} and FDST \citep{fang2019locality}.

\begin{table}[htbp]
\centering
\caption{Main statistic information on the six adopted datasets}
\scalebox{0.8}{
\begin{tabular}{ccccccc}
\toprule
Dataset & Set Count & Avg. Count & Avg. Resolution & Train & Validation & Test \\
\midrule
SHHA & 241,677 & 501 & 589*868 & 270 & 30 & 182\\
SHHB & 88,488 & 123 & 768*1024 & 360 & 40 & 316\\
QNRF & 1,251,642 & 815 & 2013*2902 & 961 & 240 & 334\\
JHU & 1,515,005 & 346 & 1430*910 & 2,772  &500&1,600 \\
NWPU & 2,133,375 & 418 & 2191*3209 & 3,109&500 &1,500 \\
FDST & 394,081 & 27 & 1080*1920 & 7,800 &1,200&6,000 \\
\bottomrule
\end{tabular}}
\label{data_information}
\end{table}

\begin{figure}[hbpt]
    \centering
    \includegraphics[width=0.8\textwidth]{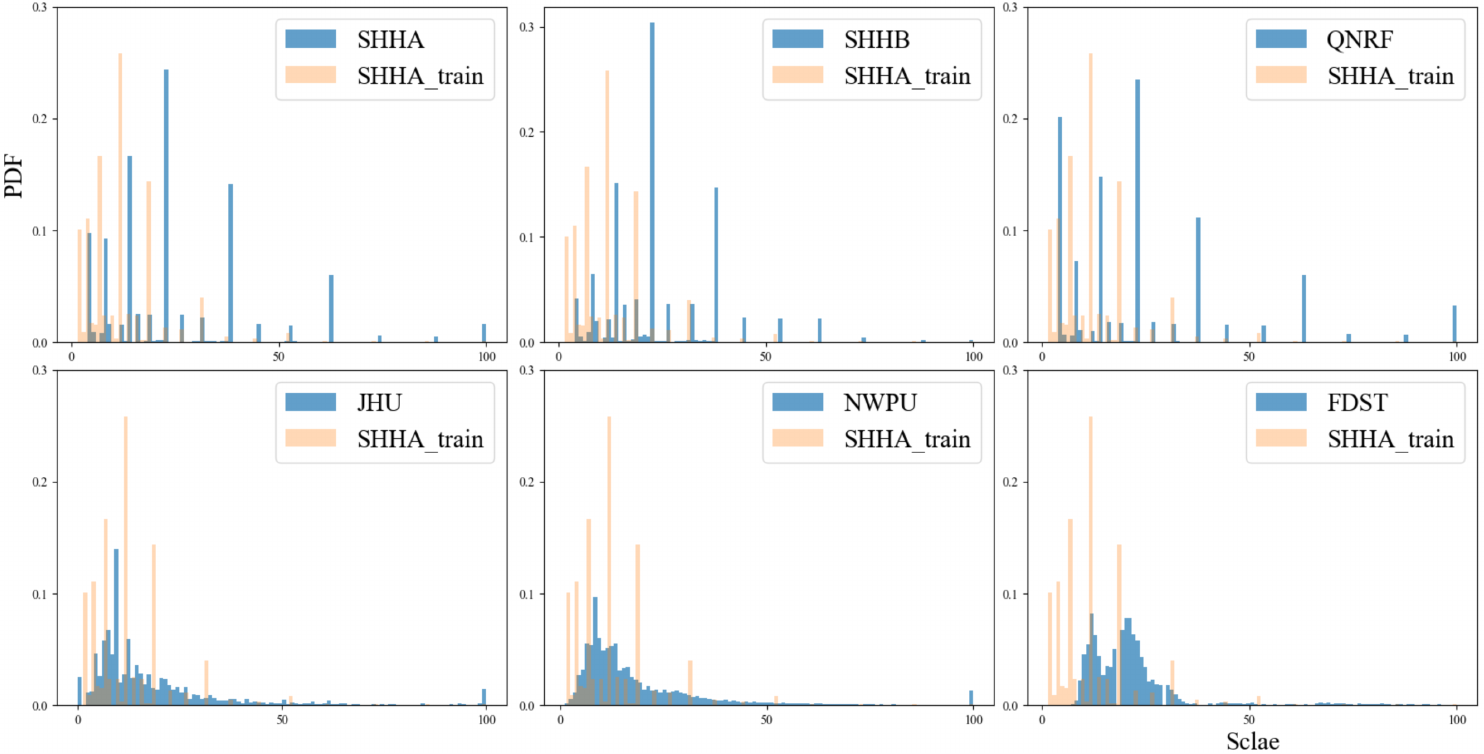}
    \caption{Scale distribution comparison between SHHA with other
adopted datasets.}
    \label{fig11}
\end{figure}

To further show the main statistic information and the domain shift existing among them, we provide some main features of the datasets in table \ref{data_information}. 
We pick some explicit domain specific knowledge to array. For the RGB images, the pixel values distribution is one of the domain specific knowledge, due to the RGB distribution representing the scene style. 
As Fig. \ref{fig11} shown, SHHA owns clear distribution with other datasets. Then, for crowd scenes, the resolution level, congested level and scale level also make great influence to the convergence. 
Hence, we show that despite that the counting of SHHA is not least, considering its average resolution, the quality of SHHA is worst. 
Finally, we calculate the annotated boxes area as the scale information to the instances and present the scale shift to datasets as shown in Appendix Fig. \hyperref[fig12]{C.1}. 
To this end, we pick the \textbf{weakest} dataset, which means least information, SHHA as our source domain, while other datasets are adopted as target domain.

\subsection{Implementation Details}
In the training phase, the training data only comes from SHHA and the model is tested on the target sets. For code backgrounds, we leverage a PyTorch framework of C3F\cite{gao2019c} on an NVIDIA A100 GPU with a memory of 80Gb. For data preparing, we randomly crop the original images with a resolution of (512 × 512), then an augmentation of random rescale with a range of [0.8, 1.2] and a probability of 0.5 for horizontal flip are leveraged. For network, a backbone model of VGG-16 \cite{simonyan2014very} and Feature Pyramid Network(FPN) \cite{lin2017feature} are adopted. For training, a batch size of 8, an optimization of Adam along with a learning rate of 1e-5 are utilized. To measure the performance on crowd localization, we utilize F1-measure (F1), precision(Pre.) and recall(Rec.), in which the F1 is the primary metric.
\begin{equation}
    \begin{aligned}
        \text{Pre.} &= \frac{TP}{TP + FP}, \\
        \text{Rec.} &= \frac{TP}{TP + FN}, \\
        \text{F1} &= \frac{2 \cdot \text{Pre.} \cdot \text{Rec.}}{\text{Pre.} + \text{Rec.}}.
    \end{aligned}
\end{equation}

\subsection{Discussion on Our Method}
\subsubsection{ Comparison between DPD and IIM}
\begin{figure*}[ht]
    \centering
    \begin{minipage}{0.5\textwidth}
        \centering
        \includegraphics[width=\textwidth]{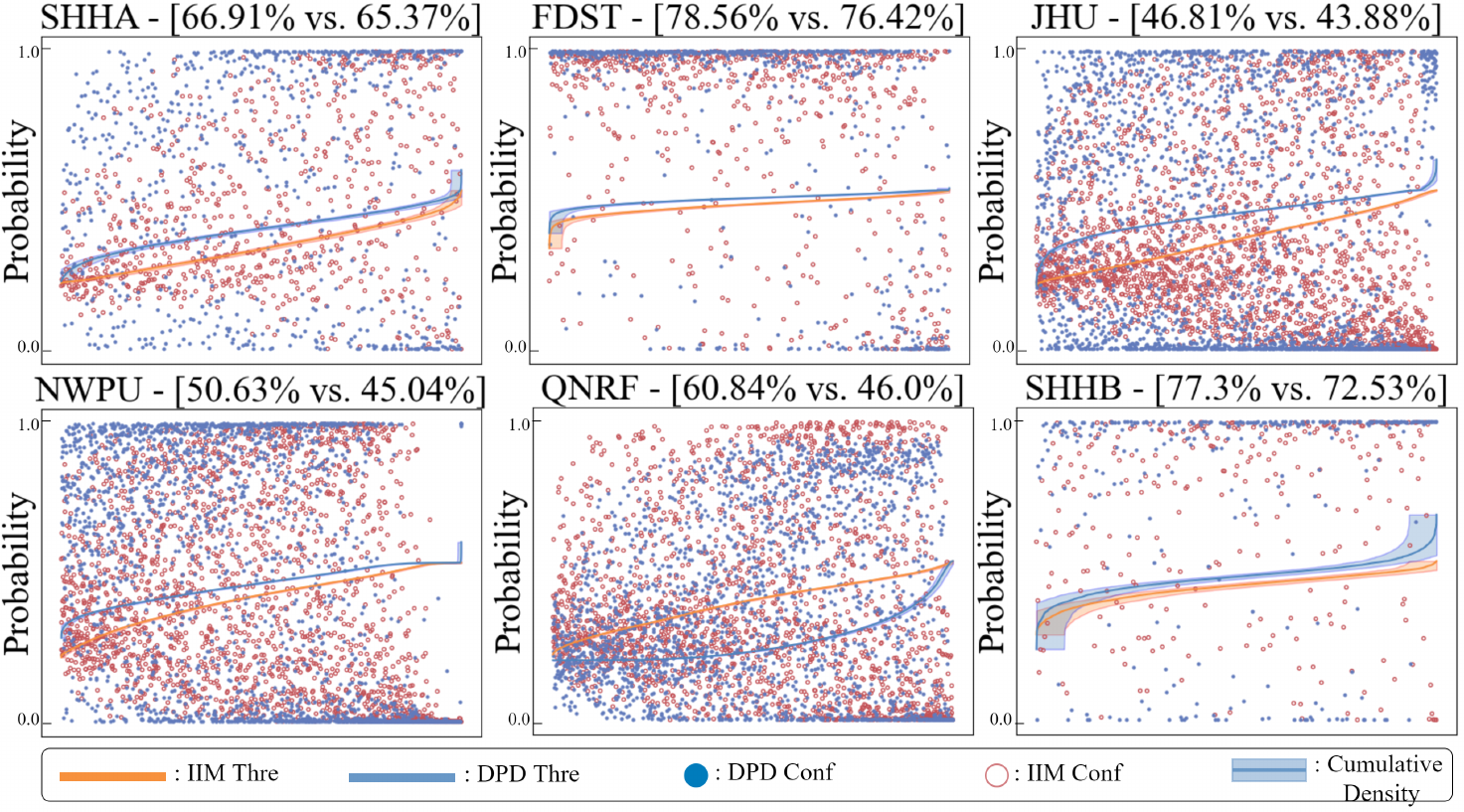}
        \caption{ 
        The confidence and threshold distribution on six adopted datasets with IIM along proposed DPD.
        The scatters in the figure are the confidences, while the plots are the thresholds, in which the shadow area denotes the density of the values, the bigger of the shadow areas are, the lower of the density is. The compared ratio is the confidence larger than its threshold.}
        \label{fig5}
    \end{minipage}\hfill
    \begin{minipage}{0.48\textwidth}
        \centering
        \includegraphics[width=\textwidth]{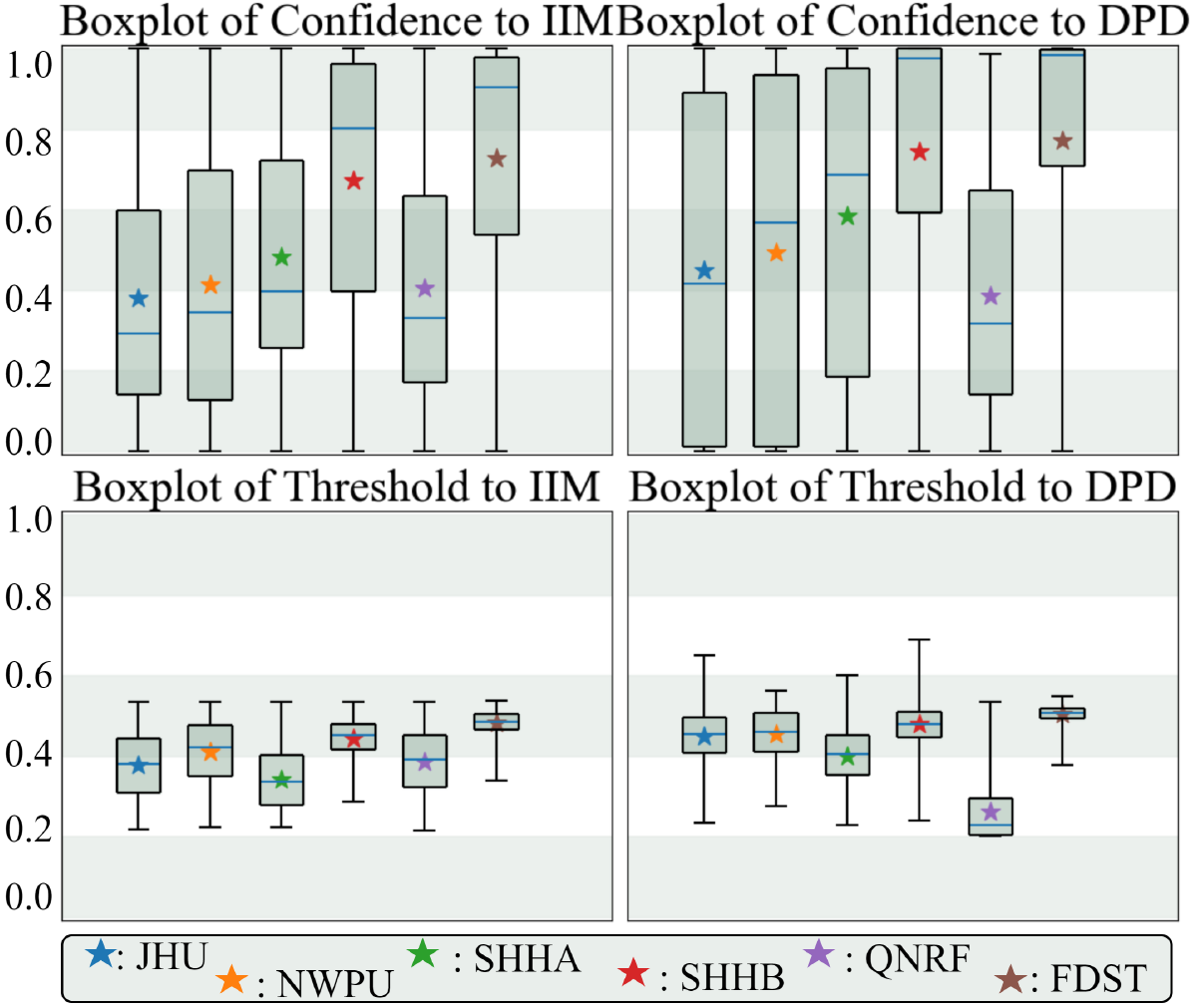}
        \caption{The boxplot of confidences (upper row) / thresholds (lower row) distribution between the IIM with DPD, in which the line in the box denotes the median and the star denotes the mean value to the distribution.}
        \label{fig6}
    \end{minipage}
\end{figure*}

In this part, we visualize the confidence and threshold distribution of the positive pixels for DPD and IIM\citep{gao2020learning}. To be concrete, our motivation is from the irrational distribution between confidence and threshold. The irrationality comes from two perspectives. 
1) The threshold is not generalized and only limited within a small value band (see Fig. \ref{fig5} \& \ref{fig6}). 
2) The uncertainty of confidence is large, which is incurred by under-fitting to target domain(see Table. \ref{MCU}).

\begin{table}[hbpt]
\small
\centering
\caption{The \textit{Monte Carlo Uncertainty} values are arrayed when the datasets are adopted as the target domains. The bold values represent better results.}
\begin{tabular}{c|c|c|c|c|c} 
\hline
\multirow{2}{*}{Datasets} & \multicolumn{5}{c}{Monte Carlo Uncertainty$^\downarrow $}                                         \\ 
\cline{2-6}
                          & JHU            & SHHB           & FDST           & QNRF           & NWPU            \\ 
\hline
Adaptive                  & 0.359          & 0.358          & 0.351          & 0.359          & 0.357           \\ 
\hline
DPD                       & \textbf{0.123} & \textbf{0.069} & \textbf{0.062} & \textbf{0.238} & \textbf{0.113}  \\ 
\hline
\end{tabular}
\label{MCU}
\end{table}

Then, we make further accurate and fastidious analysis. As shown in Fig. \ref{fig6}, we array the boxplot of the confidences and thresholds distributions on six target domains (including SHHA-test). To be concrete, the upper and lower quartile, median and extremum values are exhibited. According to the Fig. \ref{fig6}, we notice some general phenomenon. 1) The range of thresholds in DPD is expanded comparing with IIM; 2) The compactness of the DPD distribution is enhanced; 3) The average thresholds are improved except for QNRF. Then, we make discussion on the three aforementioned phenomenon.

For 1), a wider range of thresholds are obtained by introducing DPD. It is obvious that our DPD endows the threshold learner more tolerance to the outliers. As for 2), the convergence to the non-convex loss landscape is inclined to overfit on the normal samples. However, the thresholds arrayed in Fig. \ref{fig6} are all measured as target domain, which means the introduce of DPD indeed enhances the generalization. Considering 3), the QNRF dataset is extremely congested, which means it is more obscure than the source domain. Also, as the Fig. \ref{fig5} and \ref{fig6} shown, the average confidences on the QNRF is the lowest comparing with other datasets. Hence, to adapt to the difficulty of QNRF, our DPD pushes the thresholds towards 0. Then, for other datasets, the improvement can be similar. We also provide visualization results in Appendix Fig. \hyperref[fig13]{D.1} for both methods.

Besides, to measure how the uncertainty changes after introducing DPD, we compute them directly. The computation process is based on the \textit{Monte Carlo Uncertainty}:
\begin{equation}
    \mathcal{U}_{MCU}=-\frac{1}{N}\sum_{i=1}^{N} conf _i\cdot \log (conf _i),
\end{equation}

\subsubsection{Comparison between Fixed 0.3 and 0.5}
\begin{figure*}[htbp]
    \centering
    \begin{minipage}{0.5\textwidth}
        \centering
        \includegraphics[width=\textwidth]{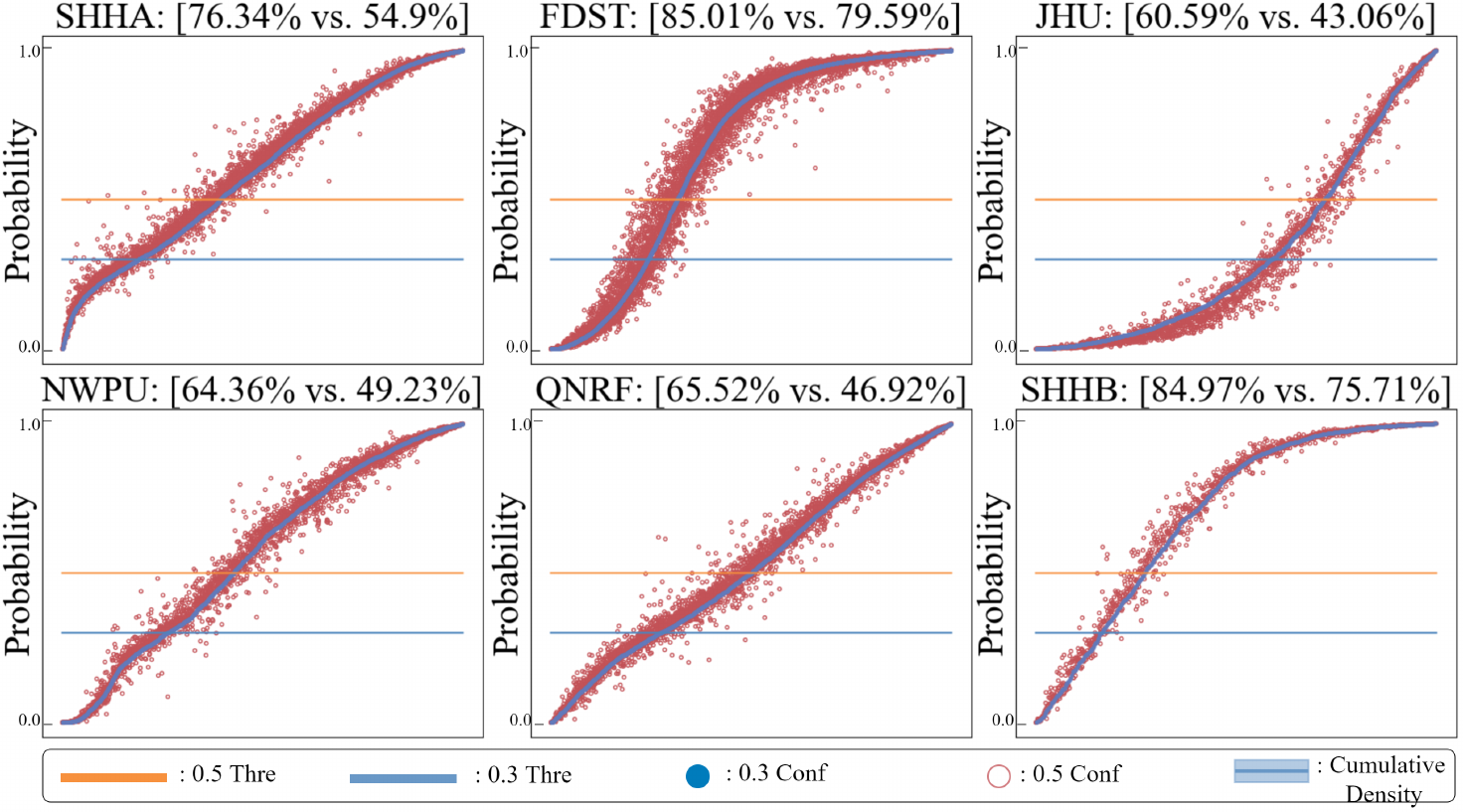}
        \caption{ 
        The confidence distribution to the crowd locator trained with the fixed thresholds, namely 0.3 and 0.5.}
        \label{fig7}
    \end{minipage}\hfill
    \begin{minipage}{0.48\textwidth}
        \centering
        \includegraphics[width=\textwidth]{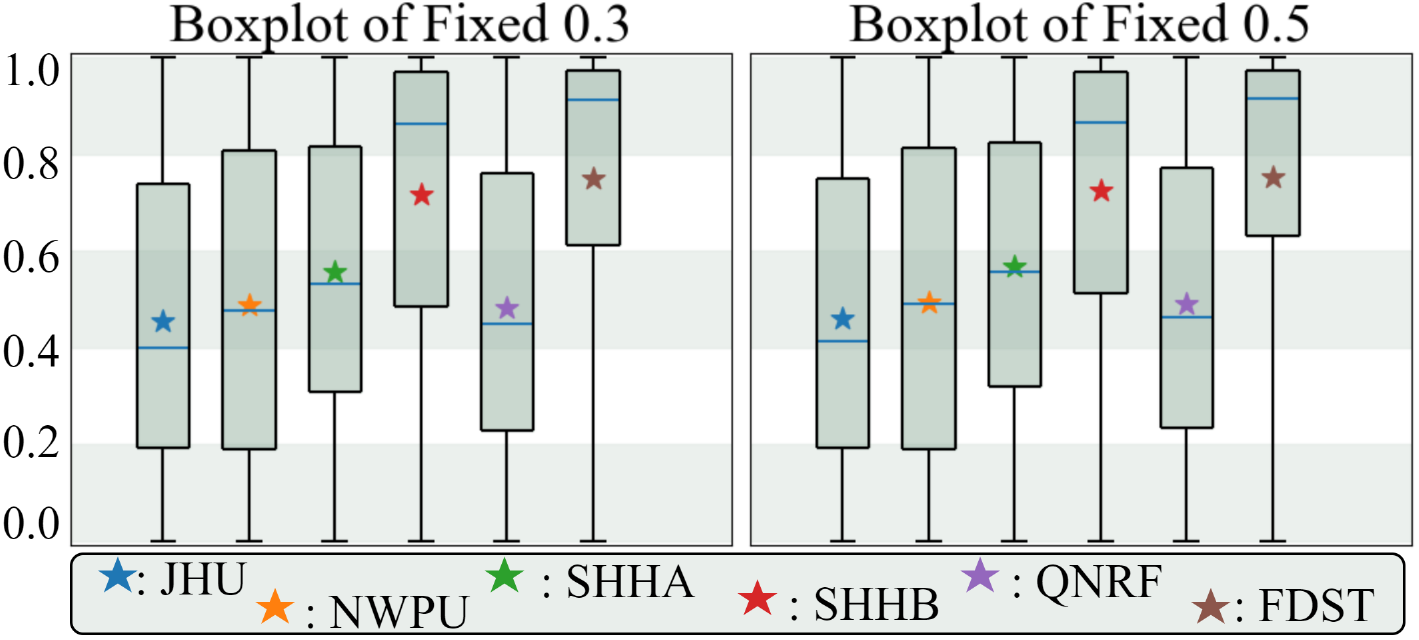}
        \caption{The boxplots of six adopted datasets, in which the left one is from IIM results, while the right one is from our proposed DPD results. The upper row is confidence, while lower row is threshold.}
        \label{fig8}
    \end{minipage}
\end{figure*}

In this subsection, we compare the crowd locators trained with fixed threshold, namely 0.3 and 0.5. Concretely, we notice that the training paradigm between adaptive threshold with fixed threshold are different. In fixed threshold training, there is no binary restraints, which means the confidence predictor is inclined to show higher confidence to positive samples (including negative samples). To this end, we show that the model selection under different threshold influences final results. As shown in the Fig. \ref{fig7} and \ref{fig8}, the higher thresholds are distributed with more uncertainty. This is because the higher threshold leaves the confidence predictor more tolerance and variance. Therefore, the scopes of two thresholds are similar but the variance are with difference.

\subsection{Experimental Guarantees}
\subsubsection{Convergence to DPD}
\begin{figure}[hbpt]
    \centering
    \includegraphics[width=0.9\textwidth]{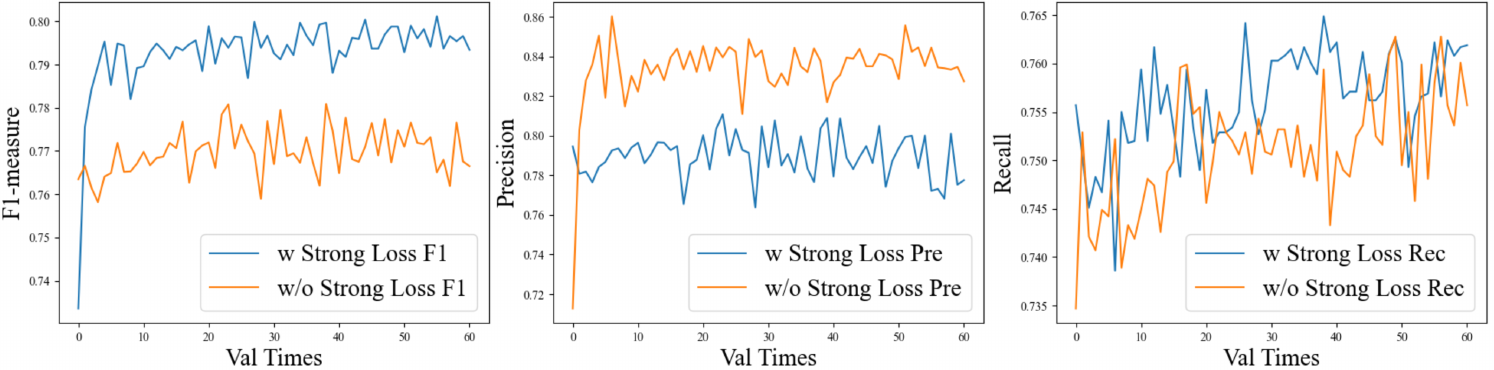}
    \caption{The training curve between \textit{w.} and \textit{wo.} strong loss.}
    \label{fig9}
\end{figure}

we provide some empirical guarantees on the convergence strategy to the dynamic proxy domain. Recall that the proposed strategy suggests a stronger loss function (Eq. \ref{eq24}), which means introduces more gradient optimization, adopted in dynamic proxy domain convergence aids crowd locator generalize well. To demonstrate the proposition empirically, we compare the convergence process under two settings. Specifically, we visualize the training curve between two settings, namely with and without strong loss in the main text, which is as shown in Fig. \ref{fig9}.

Concretely, the model trained with strong loss is the strategy arrayed in the main text, while the model without strong loss means only $\mathcal{L}_1$ loss is adopted. As shown in the Fig. \ref{fig9}, the model with stronger loss converges faster and keeps stable in the high performance. To this end, the stronger loss is indeed helpful in converging to the dynamic proxy domain and learning with more generalization.

\subsubsection{Influence Number of Samples in Gradient Descend}
In this subsection, we investigate how the number of training samples influences the final results. Theoretically, we prove that more samples within a ERM training introduces better generalization. However, in real convergence, a feasible way to improve number of samples is to enhance the batch size, which is a hyper parameters in optimization process. To this end,
the issue lies in the point. In the main text, we utilize a momentum updated model to implement the first order dynamic proxy domain optimization and alleviate the issue simultaneously. Therefore, we make further analysis from the aspect of empirical results.

\begin{figure}[hbpt]
    \centering
    \includegraphics[width=0.7\textwidth]{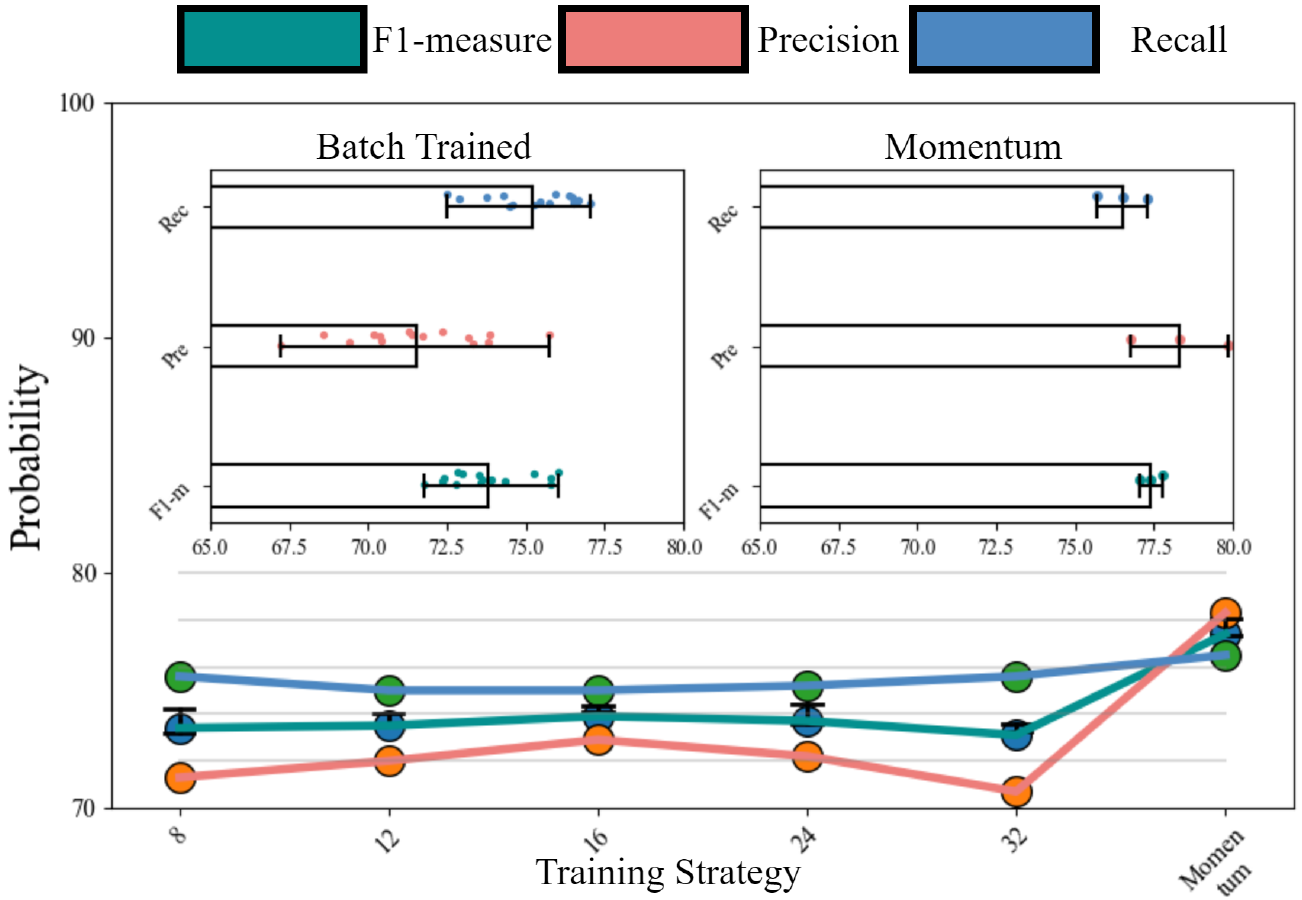}
    \caption{Comparison of models trained under different batch size and momentum manner. We pick three trained models under one setting.}
    \label{fig10}
\end{figure}

As shown in the Fig. \ref{fig10}, the curve denotes variance of metrics namely F1-measure, precision and recall for models trained under different batch size or strategy. For every setting, we pick three models, then visualize median, minimum and maximum on the boxplots. To be concrete, when the batch size is improved under tipping point, it indeed facilitates generalization. Nevertheless, a larger batch size introduces worse generalization. However, our adopted momentum training strategy only adopted number of samples which is two times than normal, which is 8 in our baseline, but we achieve a best generalization.

\subsubsection{Order for Generating DPD}
In this subsection, we empirically demonstrate the proposals shown in Thm. \ref{thm:number} respectively. The results are arrayed in Table. \ref{ablation}. As shown in Table. \ref{ablation}, each component based on our theoretical guarantees also has consistent experimental guarantee.
As shown, we can treat the \textit{Multi-Crop Momentum Training} as the combination of the first order generated dynamic proxy domain with the improved number of samples, while the \textit{Dynamic Proxy Domain} can be treat as the second order dynamic proxy domain. It is obvious that once the generation manner owns a higher order, a better result is obtained. 
\begin{table}[htbp]
\centering
\small
\caption{Experimental guarantees on our proposal. Each component is  corresponding to a theoretical guarantee.}
\begin{tabular}{c|c|c|c} 
\hline
Guarantees                                                             & F1 (\%)          & Pre.   (\%)         & Rec.  (\%)           \\ 
\hline
\begin{tabular}[c]{@{}c@{}}Baseline Zero Order\end{tabular}                                                               & 73.4 \scriptsize{$\pm$0.55}           & 71.3 \scriptsize{$\pm$0.15}          & 75.6 \scriptsize{$\pm$0.25}           \\ 
\hline
\begin{tabular}[c]{@{}c@{}}Multi-Crop Momentum Training\end{tabular} & 77.4\text{ }\scriptsize{$\pm$0.35} & 78.3\text{ }\scriptsize{$\pm$1.55} & \textbf{76.5\text{ }\scriptsize{$\pm$0.80}}  \\ 
\hline
\begin{tabular}[c]{@{}c@{}}Dynamic Proxy Domain\end{tabular}         & \textbf{80.1\text{ }\scriptsize{$\pm$0.10}} & \textbf{85.0\text{ }\scriptsize{$\pm$0.60}} & 75.7\text{ }\scriptsize{$\pm$0.50}  \\
\hline
\end{tabular}
\label{ablation}
\end{table}

\subsubsection{Generating DPD in Different Manner}

\begin{table}[hbpt]
\small
\centering
\setlength{\tabcolsep}{3mm}
\caption{Comparison on DPD generation manner. The \textit{zero order} denotes the source domain convergence. \textit{Input Gauss} and \textit{Embed Gauss} are averaged through three $\sigma$ namely $\{0.1, 0.2, 0.5\}$.}
\label{genDPD}
\begin{tabular}{c|c|c|c} 
\hline
Pertubation  & \textbf{F1 (\%)} & Pre. (\%)      & Rec. (\%)                                                 \\ 
\hline
Zero Order   & 73.4 \scriptsize{$\pm$0.55}     & 71.3 \scriptsize{$\pm$0.15}  & 75.6 \scriptsize{$\pm$0.25}                                             \\ 
\hline
MC Dropout  & 71.3 \scriptsize{$\pm$3.00}      & 68.0 \scriptsize{$\pm$4.60}  & 75.1 \scriptsize{$\pm$1.00}                                             \\ 
\hline
Color Jitter & 74.3 \scriptsize{$\pm$0.05}      & 72.3 \scriptsize{$\pm$0.30}   & \textbf{76.3 \scriptsize{$\pm$0.40}   }                                          \\ 
\hline
Embed Gauss  & 73.6 \scriptsize{$\pm$0.55}      & 71.1 \scriptsize{$\pm$1.20}  & \textbf{76.3 \scriptsize{$\pm$0.20}  }                                           \\ 

\hline
Input Gauss  & 75.3 \scriptsize{$\pm$0.25}      & 74.4 \scriptsize{$\pm$0.25}  & 76.2 \scriptsize{$\pm$0.75}                                             \\ 
\hline
Ours DPD     & \textbf{80.1 \scriptsize{$\pm$0.10}}      &  \textbf{85.0 \scriptsize{$\pm$0.60}}   & \begin{tabular}[c]{@{}c@{}} 75.7 \scriptsize{$\pm$0.50}  \\\end{tabular}  \\
\hline
\end{tabular}
\end{table}

In this subsection, we exploit some other manners to generate the dynamic proxy domain. To begin with, we deem the following manners could also expand the training domain, in which the results are shown in Table. \ref{genDPD}. Intuitively, data augmentation is an usual method. However, in pixel-wise scene understanding, the to keep augmented images being consistency, only the color-jitter is implemented. What' s more, some pertubations could also have a probability on changing the data distribution. We conduct two kinds of pertubations: (1) input pertubation, in which the input images are conducted with Gaussian noise, more specifically, we conduct three levels noise ratios; (2) intermediate pertubation, in which the intermediate representations to the images are conducted Gaussian noise, in which we conduct three levels ratios; (3) model pertubation, in which the model representation are conducted with Monte Carlo Dropout to simulate the model pertubation.

\subsection{Main Result}

\begin{table*}[hbpt]
\centering
\caption{The main results of the crowd localization under five kinds of domain generalization settings. All results are repeated three times.}
\scalebox{0.54}
{
\begin{tabular}{c|ccc|ccc|ccc|ccc|ccc}

\hline

\multirow{2}{*}{Method} & \multicolumn{3}{c|}{SHHA to SHHB (\%)} & \multicolumn{3}{c|}{SHHA to QNRF (\%)} & \multicolumn{3}{c|}{SHHA to JHU (\%)} & \multicolumn{3}{c|}{SHHA to NWPU (\%)} & \multicolumn{3}{c}{SHHA to FDST (\%)} \\ \cline{2-16}
 & \textbf{F1} & Pre. & Rec. & \textbf{F1} & Pre. & Rec. & \textbf{F1} & Pre. & Rec. & \textbf{F1} & Pre. & Rec. & \textbf{F1} & Pre. & Rec. \\ \hline

\hline Thre. 3 & 70.20 & 66.53 & \textbf{74.29} 
               & 60.64 & 66.57 & \textbf{56.21} 
               & 50.14 & 57.24 & {44.60} 
               & 58.57 & 61.73 & \textbf{55.72} 
               & 34.38 & 21.52 & \textbf{85.47} \\
\hline Thre. 5 & 74.69 & 78.44 & 71.29 
               & 60.01 & 77.55 & 48.94 
               & 50.86 & {72.11} & 39.28 
               & 60.01 & {75.47} & 49.81 
               & 64.51 & {54.51} & 79.00 \\
\hline RSC     & 73.92 & 74.75 & 73.11 
               & 59.23 & 69.00 & 51.88 
               & 50.77 & 62.18 & 42.90 
               & 58.11 & 65.62 & 52.14 
               & 33.71 & 21.20 & 82.06 \\
\hline EFDM    & 75.65 & 78.57 & 72.94 
               & 60.02 & 74.02 & 50.47 
               & 51.34 & 66.50 & 41.80 
               & 60.21 & 73.29 & 51.08 
               & 48.38 & 34.51 & 80.90 \\
\hline IRM     & 75.23 & 76.28 & \underline{74.22} 
               & 61.80 & 71.31 & \underline{54.53} 
               & 53.23 & 63.99 & \textbf{45.57} 
               & 60.18 & 67.93 & \underline{54.01} 
               & 40.09 & 26.55 & \underline{81.88} \\
\hline CORAL   & 76.43 & 78.87 & 74.14 
               & 62.45 & 74.39 & 53.81 
               & 54.38 & 68.31 & \underline{45.17} 
               & {61.98} & 74.26 & 53.19 
               & 60.81   & 48.37 & 81.87 \\
\hline IIM     & 75.57 & 78.33  & 73.00 
               & 61.13 & 74.38  & 51.88 
               & 51.76 & 66.83  & 42.24 
               & 61.08 & 74.92  & 51.55 
               & 62.44 & 51.33  & 79.67 \\
\hline OT-M    & 76.89 & 81.82  & 72.52  
               & 62.37 & 76.94  & 52.43
               & 53.46 &\textbf{72.21}  & 42.44
               & 62.28 & 77.65 & 51.98
               & \underline{67.48} & \underline{58.33} & 80.03\\
\hline STEERER & \underline{77.45} & \underline{83.24} & 72.41 
               & \textbf{63.44}    & \underline{79.97} & 52.57
               & \underline{54.47} & 71.63 & 43.94
               & \underline{63.79} & \underline{79.63} & 53.20 
               & 66.22 & 56.17 &  80.64\\
\hline DPD     & \textbf{78.61} & \textbf{84.28} & 73.66 
               & \underline{63.35} & \textbf{80.62} & 52.17 
               & \textbf{54.63} & \underline{71.89} & 44.09 
               & \textbf{64.24} & \textbf{79.97} & 53.68 
               & \textbf{68.98} & \textbf{60.83} & 79.66 \\
\hline
\end{tabular}}
\label{main}
\end{table*}

In this subsection, we compare our proposed DPD with some crowd locators adopting different thresholds. To be concrete, we set thresholds at 0.3, 0.5 and select methods like IIM \citep{gao2020learning}, RSC \citep{huang2020self}, EFDM \citep{zhang2022exact}, IRM \citep{arjovsky2019invariant}, CORAL \citep{sun2016deep}, OT-M \citep{lin2023optimal} and STEERER \citep{han2023steerer} for better comparisons. We notice that DPD performs well under most circumstances.
As shown in the Table. \ref{main}, the adopted norms are F1(\%), \textit{precision}  and \textit{recall}, in which the F1(\%) is the main metric. 

Within cross-dataset scenarios, the DPD algorithm significantly outperformed other methods, particularly when addressing the SHHA to SHHB dataset. Its superior F1 score, precision, and recall rates attest to its exceptional performance on datasets with high similarity. This performance advantage may be attributed to the robust mechanisms of DPD in feature extraction and generalization, enabling it to more effectively capture and utilize commonalities across varying scenes. However, it was observed that the recall rate of DPD was marginally lower when the target domain was the FDST dataset, especially at a threshold setting of 0.3. This phenomenon could indicate a potential over-sensitivity in predicting the number of instances, leading to an increase in false positives. Furthermore, the low threshold setting might have relaxed the criteria for instance selection, enhancing the recall rate but at the cost of precision. This trade-off reflects the necessity for finer adjustments of DPD in specific contexts. Overall, DPD demonstrated exemplary performance across multiple cross-dataset scenarios, validating its robust generalization capability in the realm of domain adaptation.

\section{Conclusion}
\label{sec: Conclusion}
In this paper, we are motivated by enhancing the generalization of crowd localization to agnostic domains. We exploit the generalization issue from the irrationally paired thresholds and confidences. To tackle the issue, we theoretically prove introducing a dynamic proxy domain deduces the generalization error risk upper bound to target domain and experimentally propose a corresponding DPD model to demonstrate the empirical effectiveness on five domain generalization settings. To the best of our knowledge, this paper firstly makes attempt on domain generalization crowd localization. We hope this study could attract more researchers' attention on the issue.  


\bibliography{mybibfile}

\FloatBarrier
\newpage

\section*{Appendices}
\FloatBarrier

\subsection*{A. Pseudo Code}
\subsubsection*{1. Data Flow for Instance Segmentation Locator.} 
\label{pseudo1}
To better clarify the pipeline of adaptive threshold crowd localization, we provide a pseudo code of data flow.
\begin{algorithm}
\caption{Data Flow for Instance Segmentation Locator}
\begin{algorithmic}[1]
\item[\textbf{Input:}] Training image $x \in \mathbb{R}^{3 \times H \times W}$, Training binary map $y^b_{gt} \in \mathbb{N}_{\{0,1\}}^{1 \times H \times W}$, Encoder $h_E$, Decoder $h_D$ and Threshold leaner $h_T$.
\item[\textbf{Output:}] Pixel-wise classification prediction $y^b_{pre}$.
\Procedure{Forward}{}
    \State Feed $x$ as input to $h_E$, then derive $h_E(x) \in \mathbb{R}^{ch \times H' \times W'}$, in which $ch \gg 3$ and $H' < H$, $W' < W$;
    \State Feed $h_E(x)$ as input to $h_D$, then derive $y^c_{pre} \in \mathbb{R}^{1 \times H \times W}$,
    \State Feed $h_E(x)$ as input to $h_T$, then derive $y^t_{pre} \in \mathbb{R}^{1 \times H \times W}$,
    \State Derive $y^b_{pre} \in \mathbb{N}_{\{0,1\}}^{1 \times H \times W}$ via $\left \lceil y^c_{pre} \ge  y^t_{pre}  \right \rfloor$.
    
\EndProcedure
\Procedure{Backward}{}
    \State Compute loss $\mathcal{L}$ in the main text;
    \State Update parameters according to $\nabla g = \frac{\partial \mathcal{L}}{\partial \theta _{\{h_E, h_D, h_T\}}}$
\EndProcedure
\end{algorithmic}
\end{algorithm}

\subsubsection*{2. Dynamic Proxy Domain Algorithm.} 
\label{pseudo2}

The pseudo code below shows our DPD training flow in detail, supplementing it with theoretical and practical inferences.

\begin{algorithm}
\caption{Dynamic Proxy Domain Algorithm}
\begin{algorithmic}[1]
\item[\textbf{Input:}] Empirical source domain $\mathcal{D}_s$; Main hypothesis mapping function $h$, Momentum hypothesis mapping function $\mathcal{M}_{M_o}$, Dynamic proxy domain generator $h^{DPD}_T$; Empirical target domain $\mathcal{D}_t$;
\Procedure{Train}{}
    \State Initialize $h$ and $\mathcal{M}_{M_o}$ with ERM on $\mathcal{D}_s$ 
    \State Initialize $h^{DPD}_T$ randomly
    \For{\# of gradient iterations}:
        \State Sample and crop $(x_i, y_i)$ and $(x_j, y_j)$ from $\mathcal{D}_s$;
        \State Leverage $h$ to predict $y^c_i, y^c_j$;
        \State Minimize $\mathcal{L}_{ERM}$ in {Eq.\,21} of the main text 
        
        \textcolor{blue}{\Comment{Empirical risk minimization}}
        \State Leverage $h$ to predict $y^c_i, y^c_j$;
        \State Leverage $\mathcal{M}_{M_o}$ to infer $y^{c}_{j(M_o)},y^{b}_{j(M_o)}$;
        \State Minimize $\mathcal{L}_{Momentum}$ in {Eq.\,22} of the main text 
        
        \textcolor{blue}{\Comment{Multi-crop momentum}}
        \State Leverage $h^{DPD}_{\mathcal{T}}$ to generate $y^b_{DPD}$ composing dynamic proxy domain $\mathcal{D}_p$,
        \State Minimize $\mathcal{L}_{DPD}$ in {Eq.\,24} of the main text. 
        
        \textcolor{blue}{\Comment{Dynamic proxy domain}}
    \EndFor
\EndProcedure

\Procedure{Test}{}
    \State Freeze the parameters of $h$;
    \For{$x^t$ sampled from $\mathcal{D}_t$}:
        \State Let $h$ predict $y^c_{pre},y^t_{pre}$ for $x^t$,
        \State Make binary prediction to obtain $y^b_{pre}$ via $\left\lceil y^c_{pre} \geq y^b_{pre} \right\rfloor$.
    \EndFor
\EndProcedure
\end{algorithmic}
\end{algorithm}

\newpage
\subsection*{B. Proof}
\subsubsection*{1. Proof to Lemma. 1}
\label{proof_lemma1}

\noindent \textbf{Lemma 1.} \textit{Assume that the $\mathcal{H}$ is a hypothesis space with a VC dimension of d and m is the number of training samples, drawn from $\mathcal{D}_s$. Given an $h \in \mathcal{H}$, the following inequality holds with a probability at least $1 - \delta $, where $\delta \in (0, 1)$}:
    \begin{align}\scriptsize
        \notag R_{\mathcal{T}}(h) \le {R}_{\mathcal{S}}(h)+\frac{1}{2} \hat{d}_{\mathcal{H}\bigtriangleup\mathcal{H} }(\mathcal{D}_{s}, \mathcal{D}_{t})  
        +4\sqrt{\frac{2dlog(2m)+log(\frac{2}{\delta})}{m} }+\lambda, \tag{B.1}
        \label{peq27}
    \end{align}
    in which 
    \begin{equation}
        \lambda=\inf_{\hat{h}\in \mathcal{H}}^{}\left [R_{\mathcal{S}}(\hat{h}) + R_{\mathcal{T}}(\hat{h})\right ]. \tag{B.2}
    \end{equation}
\textbf{\textit{Proof:}} To prove Lemma. 1, we should firstly introduce another two lemmas.

\noindent \textbf{Lemma 2.} \textit{Assume $\mathcal{H}$ is a hypothesis space with a VC dimension of d. Let $\mathcal{S}$ with $\mathcal{D}$ are empirically sampled based on i.i.d. from $\mathcal{D}_s$ with $\mathcal{D}_t$ respectively. Then we have {Eq.\,B.3} holds with a probability at least $1 - \delta $ for any $\delta \in (0, 1)$}:
\begin{equation}
    {d}_{\mathcal{H}\bigtriangleup\mathcal{H}}(\mathcal{D}_s, \mathcal{D}_t) \le \hat{d}_{\mathcal{H}\bigtriangleup\mathcal{H} }(\mathcal{D}_s, \mathcal{D}_t)+4\sqrt{\frac{dlog(2m)+log(\frac{2}{\delta})}{m} }\tag{B.3}
    \label{peq29} 
\end{equation}

\noindent \textbf{Lemma 3.} Let $\hat{h}, h$ be any hypothesis function defined on $\mathcal{H}$, we have {Eq.\,B.4} holds.
\begin{equation}
\left| R_S({h}, h') - R_T(h, h') \right| \leq \frac{1}{2} d_{\mathcal{H} \Delta \mathcal{H}} (\mathcal{D}_s, \mathcal{D}_t). \tag{B.4}
\label{peq30}
\end{equation}

As for the proof to Lemma 2 and 3, please refer to \citep{ben2010theory}. Finally, we are ready to prove Lemma. 1. 
\begin{align}
\notag R_T(h) &\leq R_T(\hat{h}) + R_T(\hat{h}, h) \\
\notag &\leq \left| R_T(h, \hat{h}) - R_S(h,\hat{h}) \right| + R_T(\hat{h}) + R_S(h,\hat{h}) \\
\notag &\leq R_T(\hat{h}) + R_S(h,\hat{h}) + \frac{1}{2} d_{\mathcal{H} \Delta \mathcal{H}} (\mathcal{D}_s, \mathcal{D}_t) \\
\notag &\leq R_T(\hat{h}) + R_S(h) + R_S(\hat{h}) + \frac{1}{2} d_{\mathcal{H} \Delta \mathcal{H}} (\mathcal{D}_s, \mathcal{D}_t) \\
\notag &\leq R_S(h) + \frac{1}{2} \hat{d}_{\mathcal{H} \Delta \mathcal{H}} (\mathcal{D}_s, \mathcal{D}_t) \\
&+ 4\sqrt{\frac{2d \log(2m) + \log(\frac{2}{\delta})}{m}} + \lambda. \tag{B.5}
\end{align}

\subsubsection*{2. Proof to Influence for the Number of Training Samples}
\label{proof_number}
\noindent
\textbf{Proposition 1.} \textit{With a greater $m$ in {Eq.\,B.1}, namely $\hat{m} > m$, it facilitates deriving a tighter upper bound to generalization error risk on target domain, which is as $\sup_{h \in \mathcal{H}} R_T(h|m) < \sup_{h \in \mathcal{H}} R_T(h|\hat{m})$.}

\textbf{Proof.} Considering the item in {Eq.\,B.1} $f(m) = 4 \sqrt{\frac{2d \log(2m) + \log(\frac{2}{\delta})}{m}}$, the only variance is $m$ in the item. To prove a greater $m$ aiding to derive a lower item, we compute the monotonicity of the item to $m$. To facilitate calculation, the root sign is omitted during differentiation:
\begin{align}
\notag \frac{\partial f}{\partial m} = \frac{\frac{2d}{m} \cdot m - {2d \cdot \log(2m) - \log(\frac{2}{\delta})}}{m^2} 
= \frac{2d \cdot \left[ 1 - \log(2m) \right] - \log(\frac{2}{\delta})}{m^2}. \tag{B.6}  
\end{align}

According to {Eq.\,B.6}, the $\frac{\partial f}{\partial m}$ is obviously less than zero. To this end, the $f(m)$ monotonically decreases along $m$.

\subsubsection*{3. Proof to Theorem 1}
\label{proof_theorem1}
\noindent\textbf{Theorem 1.}
\textit{Let $h$ be the binary classifier hypothesis in the $\mathcal{H}$ with a VC-dimension of $d$ and $m_s, m_p$ are the number of source/proxy samples. Let $\mathcal{D}_{p}$ be the empirical distribution drawn \textit{i.i.d.} from the dynamic proxy domain. Then, a hyper-parameter  $\gamma \in [0,1]$ is defined, which is the convex combination rate. Thus, for any $\delta \in (0,1)$, with probability at least $1-\delta$,}
\begin{align}\scriptsize
    \notag R_{\mathcal{T}}(h) &\leq \gamma \cdot \left( \hat{R}_{\mathcal{S}}(h) + \frac{1}{2} \operatorname{div}_{\mathcal{H}\bigtriangleup \mathcal{H}}\left( \mathcal{D}_{s}, \mathcal{D}_{t} \right) \right) \\
    \notag &+ (1-\gamma) \cdot \left( \hat{R}_{\mathcal{P}}(h) + \frac{1}{2} \operatorname{div}_{\mathcal{H}\bigtriangleup \mathcal{H}}\left( \mathcal{D}_{p}, \mathcal{D}_{t} \right) \right) \\
    \notag &+ \lambda_{\gamma} + 4 \sqrt{ \frac{2d \log 2 \left( m_{s} + m_{p} \right) + \log \left( \frac{2}{\delta} \right)}{m_{s} + m_{p}} }, \tag{B.7}
            \label{bound_another}
\end{align}
\textit{in which}
\begin{align}
\notag \lambda_{\gamma} &= \inf_{{h} \in \mathcal{H}} \left[ \gamma \cdot R_{\mathcal{S}}({\hat{h}}) + (1-\gamma) \cdot R_{\mathcal{P}}({\hat{h}}) + R_{\mathcal{T}}({\hat{h}}) \right]. \tag{B.8}
\end{align} 

\textbf{Proof.} Firstly, when introducing a dynamic proxy domain $\mathcal{D}_p$ and converging the source domain $\mathcal{D}_s$ and dynamic proxy domain equals converging a new domain $\mathcal{D}_{s^*}$. Therefore, we give a new definition as {Eq.\,B.8}:

\begin{equation}
\mathcal{D}_{s^*} \triangleq \gamma \cdot \mathcal{D}_s + (1 - \gamma) \cdot \mathcal{D}_p. \tag{B.8}
\end{equation}

According to the Lemma. 1, we can rewrite the source domain into source-star domain. Then, we have {Eq.\,B.9} holds for any  $\delta \in (0, 1)$, w.p.b. at least $1 - \delta $,

\begin{equation}
R_{\mathcal{T}}(h) \leq R_{S^*}(h) + \frac{1}{2} \hat{d}_{{\mathcal{H}} \triangle \mathcal{H}}(\mathcal{D}_{S^*}, \mathcal{D}_t) + 4 \sqrt{\frac{2d \log 2 (m_s + m_p) + \log(\frac{2}{\delta})}{m_s + m_p}} + \lambda_\gamma. \tag{B.9}
\end{equation}

Recall that the proposed $\mathcal{H} \triangle \mathcal{H}$ divergence is hard to compute, thus Def. 2 introduces a proxy divergence.

\noindent
\textbf{Definition 2.}
    A proxy dataset is constructed as:
        \begin{equation}
            \mathcal{X}_{prox}=\left \{ (x_i,\lceil x_i\sim \mathcal{D}_s\rfloor) |i\in \{0,\cdots ,N_s+N_t\}\right \} . \tag{B.10}
        \end{equation}
    A proxy generalized error $\epsilon_p$ is introduced on $\mathcal{X}_{prox}$. Then, using $\mathcal{A}-$distance ($A$ is some specific part of $\mathcal{X}_{prox}$ and $\mathcal{A}$ is the set of them), the $\mathcal{H}\bigtriangleup  \mathcal{H}\text{-}$divergence can be approximated as:
    \begin{equation}
        \hat{div}_{\mathcal{H}\bigtriangleup \mathcal{H}}=2\cdot (1-2\epsilon_p)=2\sup_{A\in \mathcal{A} }^{}\left | \text{Pr}_{\mathcal{D}_s}(A)-\text{Pr}_{\mathcal{D}_t}(A) \right |. \tag{B.11}
        \label{TVD}
    \end{equation}
Thus, for the relationship between $\hat{d}_{\mathcal{H} \triangle \mathcal{H}}(\mathcal{D}_{s^*}, \mathcal{D}_t)$ with $\hat{d}_{\mathcal{H} \triangle \mathcal{H}}(\mathcal{D}_s, \mathcal{D}_t)$ can be derived in the following:
\begin{align}
\notag & \hat{d}_{\mathcal{H} \triangle \mathcal{H}}(\mathcal{D}_{s^*}, \mathcal{D}_t) 
= 2 \sup_{A \in \mathcal{A}} \left| \ {\text{Pr}_{\mathcal{D}_{s^*}}}(A) - \ \text{Pr}_{\mathcal{D}_t}(A) \right| \\
\notag 
&= 2 \sup_{A \in \mathcal{A}} \left| \gamma \cdot \left[ \ \text{Pr}_{\mathcal{D}_s}(A) - \ \text{Pr}_{\mathcal{D}_t}(A) \right] + (1 - \gamma) \cdot \left[ \ \text{Pr}_{\mathcal{D}_p}(A) - \ \text{Pr}_{\mathcal{D}_t}(A) \right] \right| 
\\
\notag 
&\leq 2 \cdot \gamma \cdot \left[ \sup_{A \in \mathcal{A}} \left| \ \text{Pr}_{\mathcal{D}_s}(A) - \ \text{Pr}_{\mathcal{D}_t}(A) \right|  + (1 - \gamma) \cdot \sup_{A \in \mathcal{A}} \left| \ \text{Pr}_{\mathcal{D}_p}(A) - \ \text{Pr}_{\mathcal{D}_t}(A) \right| \right]. \tag{B.12}
\end{align}

\noindent
Then, let us rewrite {Eq.\,B.12} into $\mathcal{H} \triangle \mathcal{H}$-divergence, which is as {Eq.\,B.13}:
\begin{equation}
d_{\mathcal{H} \triangle \mathcal{H}}(\mathcal{D}_{s^*}, \mathcal{D}_t) \leq \gamma \cdot d_{\mathcal{H} \triangle \mathcal{H}}(\mathcal{D}_s, \mathcal{D}_t) + (1 - \gamma) \cdot d_{\mathcal{H} \triangle \mathcal{H}}(\mathcal{D}_p, \mathcal{D}_t).
\tag{B.13}
\end{equation}
What’s more, recall the {Eq.\,B.8}, it is obvious that the {Eq.\,B.14} holds:
\begin{equation}
R_{\mathcal{S}^*}(h) \approx \gamma \cdot R_\mathcal{S}(h) + (1 - \gamma) \cdot R_\mathcal{P}(h).
\tag{B.14}
\end{equation}
Since the two terms in the both side of {Eq.\,B.14} are all optimized in the fully-supervised manner, we can approximate them into equal pair. By now, summarizing {Eq.\,B.13} with B.14, we have:
\begin{align}
\notag R_{\mathcal{S}^*}(h) + \frac{1}{2} d_{\mathcal{H} \triangle \mathcal{H}}(\mathcal{D}_{s^*}, \mathcal{D}_t) &\leq \gamma \cdot [R_\mathcal{S}(h) + d_{\mathcal{H} \triangle \mathcal{H}}(\mathcal{D}_s, \mathcal{D}_t)]\\ 
& + (1 - \gamma) \cdot [R_\mathcal{P}(h) + d_{\mathcal{H} \triangle \mathcal{H}}(\mathcal{D}_p, \mathcal{D}_t)].
\tag{B.15}
\end{align}

\subsubsection*{4. Proof to Theorem 2}
\label{proof_theorem2}
\noindent
\textbf{Theorem 2.}
\textit{Let  $ h_{\text{DPD}} $  be the DPD hypothesis and  $ h_{\text{ERM}} $  be the Empirical Risk Minimization hypothesis on the space of  $ \mathcal{H} $  with a VC-dimension of  $ d $. Then, for any  $ \delta \in (0,1) $ , with probability at least  $ 1-\delta $ , the {Eq.\,B.16} can be derived,}
\begin{equation}
    \sup_{h_{\text{DPD}} \in \mathcal{H}} R_{\mathcal{T}}(h_{\text{DPD}}) \leq \sup_{h_{\text{ERM}} \in \mathcal{H}} R_{\mathcal{T}}(h_{\text{ERM}}), \tag{B.16}
\end{equation}

\textbf{Proof.} With the conclusion of {Lem.\,1} and {Thm.\,1} in the main text, the proof to {Thm.\,2} could be very easy. 
To begin with, let us take the supremum apart. Firstly, as aforementioned, since the $R_\mathcal{S}(h)$ and $\gamma \cdot R_\mathcal{S}(h) + (1 - \gamma) \cdot R_{\mathcal{T}}(h)$ are all optimized in fully supervised manner, the two terms can be approximately deemed as equal. Secondly, as for the $\varepsilon(\cdot)$, the {Thm.\,1} in the main text tells comparison. To this end, the point lies on the divergence relationship, which can be proven as follows:

\begin{align*}
& d_{\mathcal{H} \triangle \mathcal{H}}(\mathcal{D}_s, \mathcal{D}_t) - [\gamma \cdot d_{\mathcal{H} \triangle \mathcal{H}}(\mathcal{D}_s, \mathcal{D}_t) + (1 - \gamma) \cdot d_{\mathcal{H} \triangle \mathcal{H}}(\mathcal{D}_p, \mathcal{D}_t)] \\
&= 2 \sup_{A \in \mathcal{A}} |\ \text{Pr}_{\mathcal{D}_s}(A) - \ \text{Pr}_{\mathcal{D}_t}(A)| - [2 \cdot \gamma \cdot \sup_{A \in \mathcal{A}} |\ \text{Pr}_{\mathcal{D}_s}(A) - \ \text{Pr}_{\mathcal{D}_t}(A)| \\
& \quad + 2 \cdot (1 - \gamma) \cdot \sup_{A \in \mathcal{A}} |\ \text{Pr}_{\mathcal{D}_p}(A) - \ \text{Pr}_{\mathcal{D}_t}(A)|] \\
&= [2 \cdot \gamma \cdot \sup_{A \in \mathcal{A}} |\ \text{Pr}_{\mathcal{D}_s}(A) - \ \text{Pr}_{\mathcal{D}_t}(A)| + 2 \cdot (1 - \gamma) \cdot 2 \sup_{A \in \mathcal{A}} |\ \text{Pr}_{\mathcal{D}_p}(A) - \ \text{Pr}_{\mathcal{D}_t}(A)|]  \\
& \quad - [2 \cdot \gamma \cdot \sup_{A \in \mathcal{A}} |\ \text{Pr}_{\mathcal{D}_s}(A) - \ \text{Pr}_{\mathcal{D}_t}(A)| + 2 \cdot (1 - \gamma) \cdot 2 \sup_{A \in \mathcal{A}} |\ \text{Pr}_{\mathcal{D}_p}(A)-\ \text{Pr}_{\mathcal{D}_t}(A)|] \\
&= 2 \cdot \gamma \cdot [\sup_{A \in \mathcal{A}} |\ \text{Pr}_{\mathcal{D}_s}(A) - \ \text{Pr}_{\mathcal{D}_t}(A)| - \sup_{A \in \mathcal{A}} |\ \text{Pr}_{\mathcal{D}_p}(A) - \ \text{Pr}_{\mathcal{D}_t}(A)|]  \\
& \quad + 2 \cdot (1 - \gamma) \cdot [\sup_{A \in \mathcal{A}} |\ \text{Pr}_{\mathcal{D}_s}(A) - \ \text{Pr}_{\mathcal{D}_t}(A)| - \sup_{A \in \mathcal{A}} |\ \text{Pr}_{\mathcal{D}_p}(A) - \ \text{Pr}_{\mathcal{D}_t}(A)|] \\
&\geq 0 \tag{B.17}
\end{align*}

\subsection*{C. Datasets} \label{appendix_dataset}

To further show the main statistic information and the domain shift existing among them, we array some main features of the datasets and the results are as follows.

\renewcommand{\thefigure}{C.1}
\begin{figure*}[hbpt]
    \centering
    \includegraphics[width=0.98\textwidth]{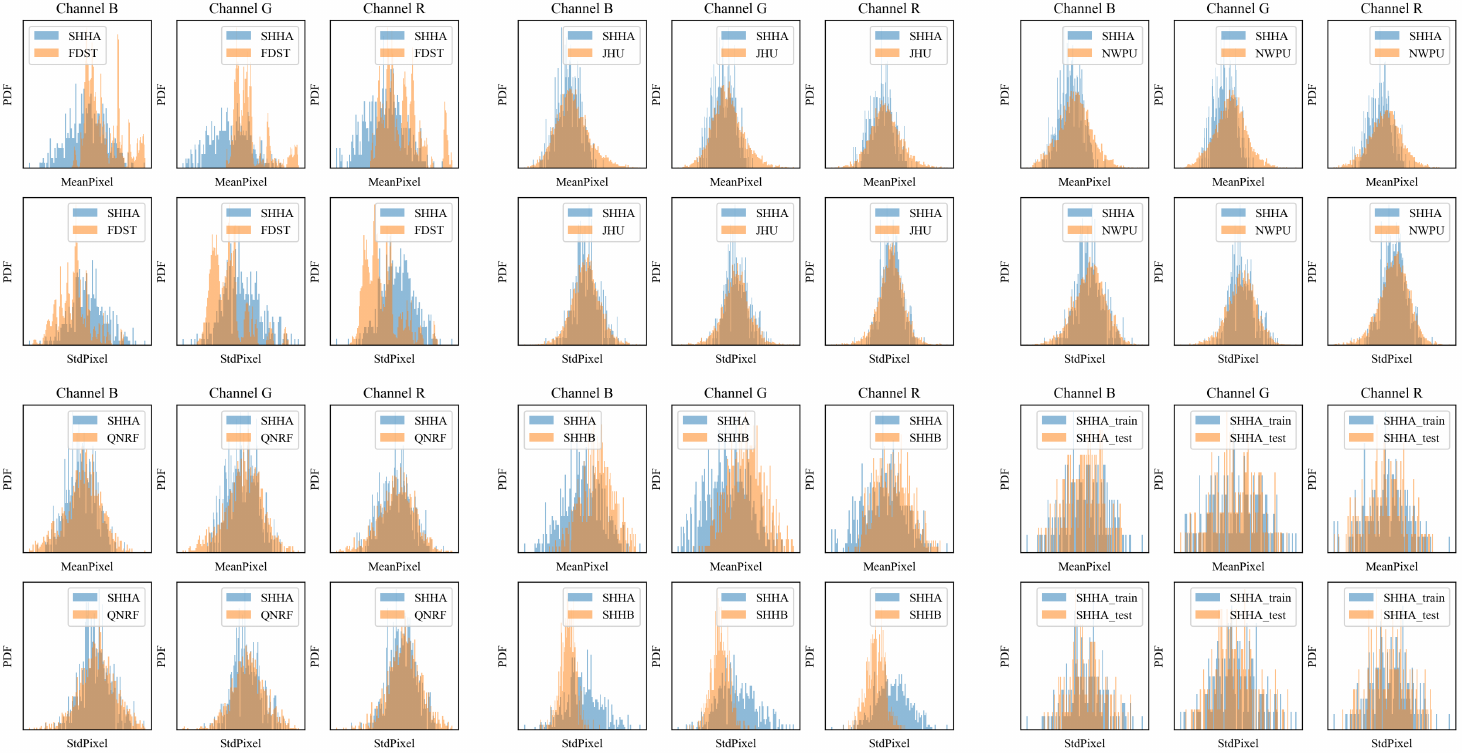}
    \caption{Scene distribution comparison between SHHA with other datasets. Concretely, the scene distribution can be decoupled into the statistics namely mean and standard deviation for pixel values in RGB channels.}
    \label{fig12}
\end{figure*}

\renewcommand{\thefigure}{D.1}
\subsection*{D. Visualization of DPD and IIM} \label{appendix_comparision}
\begin{figure*}[hbpt]
    \centering
    \includegraphics[width=0.98\textwidth]{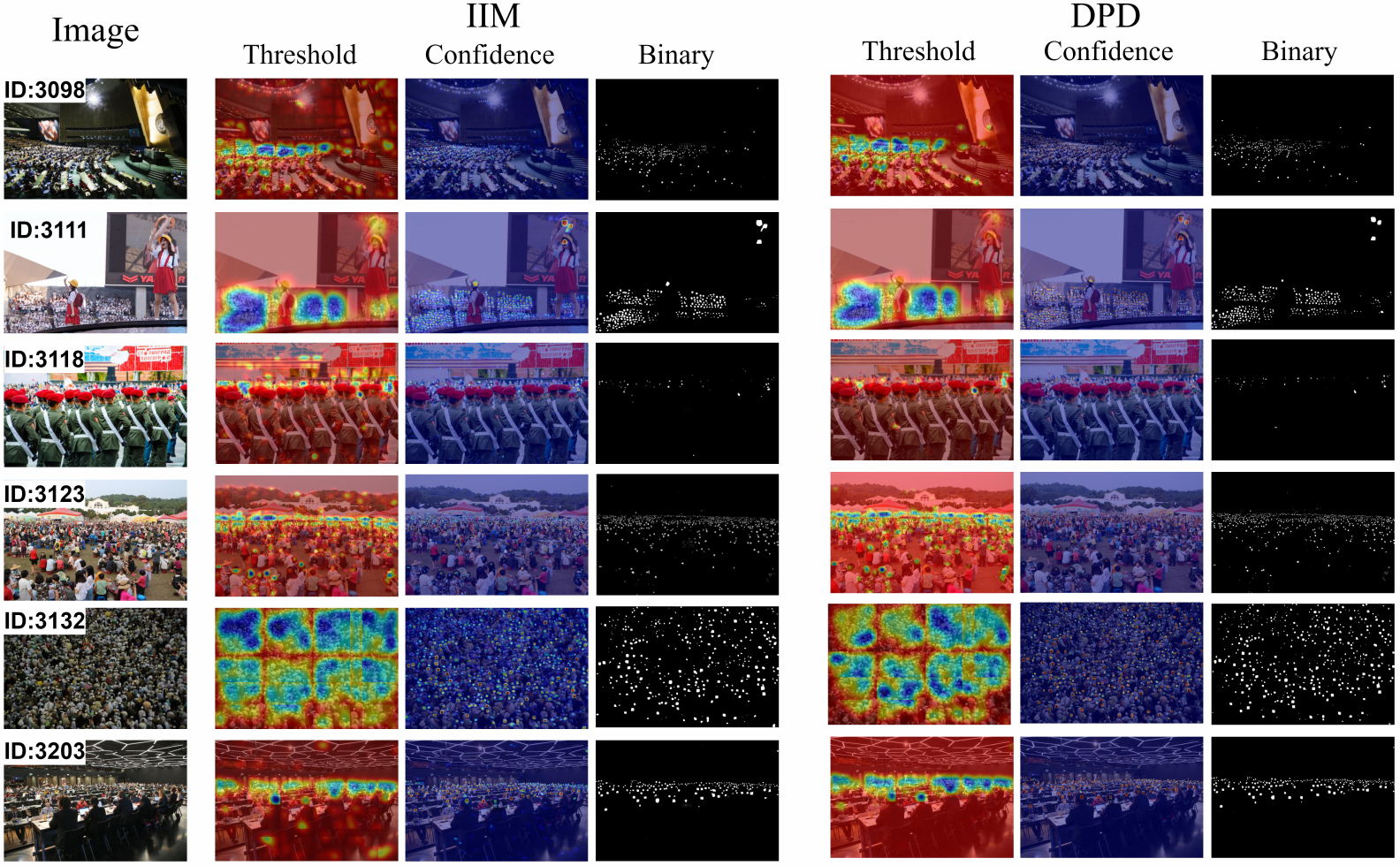}
    \caption{Some typical visualization results from NWPU-Crowd validation set.}
    \label{fig13}
\end{figure*}

\FloatBarrier
\FloatBarrier

\end{document}